\documentclass{article}
\usepackage[preprint]{log_2026}            
\makeatletter
\renewcommand{\@noticestring}{Accepted at the Fifth Learning on Graphs Conference (LoG 2026), Proceedings Track.}
\makeatother

\usepackage{setspace}
\usepackage{booktabs}
\usepackage{xcolor}
\newif\ifrevmarkup
\revmarkupfalse
\newcommand{\rev}[1]{\ifrevmarkup\begingroup\color{blue}#1\endgroup\else#1\fi}
\newcommand{\revon}{\ifrevmarkup\color{blue}\fi}
\usepackage{float}
\usepackage{wrapfig}
\usepackage{amsfonts}
\usepackage{amsmath}
\usepackage{amssymb}
\usepackage{graphicx}
\usepackage[T1]{fontenc}   
\usepackage{courier}       
\usepackage[numbers,compress,sort]{natbib}
\usepackage{tikz}
\usetikzlibrary{arrows.meta,positioning,calc,fit,backgrounds,patterns}
\usepackage{xcolor}
\usepackage{algorithm}
\usepackage{algpseudocode}
\usepackage{placeins}
\usepackage{enumitem}

\definecolor{erTeal}{HTML}{0F6E56}\definecolor{erTealBg}{HTML}{E1F5EE}
\definecolor{erGrayL}{HTML}{B8B6B0}\definecolor{erInk}{HTML}{1A1A1A}
\definecolor{erPurple}{HTML}{534AB7}\definecolor{erPurpleBg}{HTML}{EEEDFE}
\definecolor{erBlue}{HTML}{185FA5}\definecolor{erBlueBg}{HTML}{E6F1FB}
\definecolor{erGray}{HTML}{5F5E5A}\definecolor{erGrayBg}{HTML}{F1EFE8}
\definecolor{erRed}{HTML}{A32D2D}\definecolor{erRedBg}{HTML}{FCEBEB}
\definecolor{erAmber}{HTML}{9A6B14}\definecolor{erAmberBg}{HTML}{FBF0DC}
\definecolor{erCountFill}{HTML}{D9C7A6}\definecolor{erRecencyFill}{HTML}{579A89}\definecolor{erBankFill}{HTML}{3A2E6E}
\tikzset{
  erbox/.style={rounded corners=3pt, draw, line width=0.4pt, align=center,
    inner sep=4pt, font=\small, minimum height=9mm, text width=34mm},
  gray/.style ={erbox, draw=erGray,   fill=erGrayBg,   text=erGray},
  teal/.style ={erbox, draw=erTeal,   fill=erTealBg,   text=erTeal},
  purple/.style={erbox, draw=erPurple, fill=erPurpleBg, text=erPurple},
  blue/.style ={erbox, draw=erBlue,   fill=erBlueBg,   text=erBlue},
  red/.style  ={erbox, draw=erRed,    fill=erRedBg,    text=erRed},
  amber/.style={erbox, draw=erAmber,  fill=erAmberBg,  text=erAmber},
  erflow/.style={-{Stealth[length=2mm]}, line width=0.5pt, draw=erGray},
}

\title[EdgeReMIND: A Scalable, Top-Ranked Memorization Baseline]{EdgeReMIND: A Scalable, Top-Ranked Memorization Baseline for Temporal Multi-Relational Link Prediction}

\author[B.~Pollard]{%
Bryant Pollard \\
Clemson University \\
\email{bdpolla@clemson.edu} \\
\email{bryantpollardresearch@gmail.com}}

\renewcommand{\topfraction}{0.9}
\renewcommand{\bottomfraction}{0.9}
\renewcommand{\textfraction}{0.07}
\renewcommand{\floatpagefraction}{0.7}

\begin{document}

\maketitle

\begin{abstract}
Temporal link prediction on the Temporal Graph Benchmark 2.0 (TGB~2.0)~\citep{gastinger2024tgb2} faces a scalability ceiling: on the benchmark's three largest datasets, every existing embedding method runs out of memory or exceeds the time budget. These large-scale graphs are the ones nearest real deployment scale, so failing on them is a real production limitation. EdgeReMIND sets the highest reported test mean reciprocal rank (MRR) on six of eight TGB~2.0 datasets and is the only relation-aware method that runs on all of them. This linear memorization model, with learned per-relation weights over data-calibrated features, is therefore not merely a fallback where embeddings fail but a practical state-of-the-art baseline across the benchmark.
\end{abstract}

\section{Introduction}
\label{sec:introduction}

Temporal link prediction asks which future edge will form in a graph that evolves over time. This work addresses its multi-relational form: given a source node, a relation type, and a time, the task is to rank the true destination above a set of candidate destinations. It spans two principal graph types: temporal knowledge graphs (TKGs) and temporal heterogeneous graphs (THGs). TKGs link entities by typed relations, with each fact carrying a timestamp, and are applied to tasks such as medical diagnosis over evolving patient records~\citep{song2020medicaltkg}. THGs let multiple node and edge types interact over time, and are used in applications such as evolving malware detection~\citep{fan2021heterogeneous}.

The field's current frontier benchmark is TGB~2.0~\citep{gastinger2024tgb2}. That benchmark exposes a scalability ceiling that has become the binding constraint on progress. The ceiling is sharpest on the three largest datasets by node count. On these, every embedding-based method on the leaderboard is absent, out of memory or over the time budget, and no relation-aware method runs at all, leaving only pair-based EdgeBank-class memorization~\citep{poursafaei2022edgebank} (\S\ref{sec:leaderboard}). \rev{The budget is the benchmark's own: up to 40~GB of GPU memory on the evaluation hardware, and a seven-day per-experiment limit.} This gap is structural, not incidental. The per-edge memory cost of representation learning scales unfavorably with graph size~\citep{huang2023tgb,gastinger2024tgb2}, so the problem grows precisely as datasets approach deployment scale. On these datasets, raising the achievable ceiling with any method that runs within a commodity budget is valuable before absolute accuracy is even weighed.

\paragraph{Memorization heuristics are strong but limited.} Recent benchmarking finds that heavily parameterized temporal models are, on many datasets, outperformed by simple memorization heuristics~\citep{poursafaei2022edgebank,gastinger2024recb}. The most prominent is \textbf{EdgeBank}~\citep{poursafaei2022edgebank}, which predicts a link if the same pair was seen before, optionally within a recent window. Stronger variants have followed, including the training-free \textbf{Base3}~\citep{kondrup2025base3} and the per-relation \textbf{Recurrency Baseline} (RecB)~\citep{gastinger2024recb}. These methods share limitations that fall along three axes, and no prior method addresses all three. First, the fixed-rule methods (EdgeBank, Base3) are blind to relation type, scoring every query by the same rule even though some relations recur often and others almost never; RecB conditions on the relation by fitting per relation, but the next two limitations remain. Second, the balance between ``ever seen'' and ``recently seen'' is set by hand or by coarse grid search, rather than learned from the recurrence-gap statistics. Third, the recency signal uses a single decay timescale per relation, fixed in absolute time units rather than calibrated from the data's own gaps. That single timescale is mismatched both to datasets whose dynamics live at other scales and to relations within one dataset that recur at different rates.

\paragraph{Contributions.} This paper presents EdgeReMIND (\textbf{Edge} \textbf{Re}lation-aware \textbf{M}emorization with \textbf{IN}terval \textbf{D}ecay), a CPU-only, embedding-free (no learned node or entity representations) relation-aware memorization model that resolves all three limitations and scales where prior relation-aware methods cannot:

\begin{enumerate}[leftmargin=1.5em,itemsep=0.2em,topsep=0.3em]
\item \textbf{A scalable, CPU-only baseline.} EdgeReMIND is the first relation-aware temporal link-prediction model to run end-to-end on every TGB~2.0 dataset on CPU alone (a single multi-core node, no GPU), including the three largest datasets inaccessible to embedding-based methods (\S\ref{sec:scalability}).
\item \textbf{A data-calibrated multi-timescale bank.} The bank's decay rates are calibrated, unit-free, from each dataset's own training-split inter-recurrence gaps. Its zero-initialized columns give a safety property, verified per seed: the bank helps where the gap structure needs multiple timescales and is harmless by construction elsewhere (\S\ref{sec:bank}).
\item \textbf{Per-relation learning as the primary driver.} The relation both conditions the memorization features and indexes a separately learned weight vector per relation. \rev{Learning those weights, rather than applying fixed heuristic rules, is the primary driver}; EdgeReMIND thereby sets state-of-the-art test MRR on six of eight datasets (\S\ref{sec:learning}, \S\ref{sec:main-results}, \S\ref{sec:leaderboard}).
\item \textbf{A performance decomposition across components.} Per-relation learning drives the largest gains, and among the predefined memorization features recency is the most broadly load-bearing while counts become decisive on certain THGs (\S\ref{sec:learning-value}).
\end{enumerate}

\section{Related Work}

\paragraph{Memorization baselines.} EdgeReMIND draws on three lines of memorization-based work introduced in \S\ref{sec:introduction}. EdgeBank~\citep{poursafaei2022edgebank} provides the foundation that EdgeReMIND expands: its pairwise ``seen-before'' rule is the relation-agnostic source--destination recurrence that EdgeReMIND builds on and extends to relation-specific recurrence and multi-scale recency. RecB~\citep{gastinger2024recb} is another similar method, being the only one that is both per-relation and fitted to data; the decisive difference is \emph{how} it is fitted. RecB selected two recurrency parameters per relation (a single decay rate and a strict-versus-relaxed mixing weight) by validation grid search. EdgeReMIND learns a full $15$-dimensional weight vector per relation by gradient descent (\S\ref{app:parambudget}). The higher-dimensional weighting is what gives EdgeReMIND its added flexibility: gradient learning over the broader feature set lets each relation find its own balance of count, recency, and timescale signals, a balance that becomes expressible once the fit moves beyond two parameters. RecB also fixed a single decay timescale, while EdgeReMIND supplies a calibrated multi-timescale bank. Base3~\citep{kondrup2025base3}, closest to the base portion of EdgeReMIND, fused several heuristics into a training-free score with fixed interpolation weights on the single-relation datasets of the original TGB~\citep{huang2023tgb}. EdgeReMIND keeps the fuse-the-heuristics philosophy but targets the multi-relational setting, replaces Base3's dataset-global interpolation with per-relation learned weights, and adds a data-calibrated multi-timescale bank per scope.

\paragraph{Multi-timescale temporal modeling.} Multiple decay rates have a long history in temporal modeling, where mixtures of exponentials capture effects at different timescales. In neural temporal graph learning, time2vec~\citep{kazemi2019time2vec} and the time encodings used in TGAT~\citep{xu2020tgat} and TGN~\citep{rossi2020tgn} can be viewed as learning multiple time-frequency components, though as parameters of a deep network rather than as features of a memorization model. EdgeReMIND's bank is distinct in two respects. It operates inside an embedding-free model over memorization features. Its decay rates are not learned but calibrated from the data's own inter-recurrence gap statistics (\S\ref{app:calibration}), making them interpretable per-scope half-lives in the dataset's native timestamp units.

\paragraph{Learned methods.} The dominant approaches learn node and edge representations on dynamic graphs. They include embedding-trajectory methods such as JODIE~\citep{kumar2019jodie}, memory-based message passing such as TGN~\citep{rossi2020tgn} and its edge-type variants, transformer-style architectures such as STHN~\citep{li2023sthn}, and autoregressive knowledge-graph models such as RE-GCN~\citep{li2021regcn} and CEN~\citep{li2022cen}. Like these methods, EdgeReMIND learns its weights from data rather than applying a fixed rule. Unlike them, it learns only a single linear layer, 15 weights per relation, over predefined memorization features rather than a representation.

\section{Method}

\begin{figure}[ht]
\centering
\resizebox{\linewidth}{!}{%
\begin{tikzpicture}[
  node distance=4mm and 8mm,
  pbox/.style={erbox, text width=26mm, minimum height=14mm, font=\small, inner sep=4pt},
]
\node[pbox, gray] (stream)
  {{\bfseries Training stream}\\[1pt]{\scriptsize $(s,r,d,t)$}};
\node[pbox, purple, right=9mm of stream, yshift=8mm] (base)
  {{\bfseries Base extraction}\\[1pt]{\scriptsize 6 base feats, $t'<t$}};
\node[pbox, teal, right=9mm of stream, yshift=-8mm] (cal)
  {{\bfseries Bank calibration}\\[1pt]{\scriptsize $m\cdot\{\tfrac12,1,2\}$, per scope}};
\node[pbox, blue, right=9mm of base, yshift=-8mm] (learn)
  {{\bfseries Per-relation learning}\\[1pt]{\scriptsize 15-dim vec; $\theta\!:\!|\mathcal{R}|\!\times\!15$}};
\node[pbox, amber, right=8mm of learn] (test)
  {{\bfseries Epoch selection \& test}\\[1pt]{\scriptsize val-best-epoch; $f=\theta_r^\top\phi$}};
\draw[erflow] (stream.east) -- (base.west);
\draw[erflow] (stream.east) -- (cal.west);
\draw[erflow] (base.east) -- (learn.west);
\draw[erflow] (cal.east) -- (learn.west);
\draw[erflow] (learn) -- (test);
\end{tikzpicture}%
}
\caption{EdgeReMIND end-to-end (left to right): the training stream feeds base extraction and bank calibration in parallel, forming the 15-dimensional per-candidate vector that drives per-relation learning, validation-best-epoch selection, and test scoring. Bank calibration is expanded in Figure~\ref{fig:calib}.}
\label{fig:pipeline}
\end{figure}
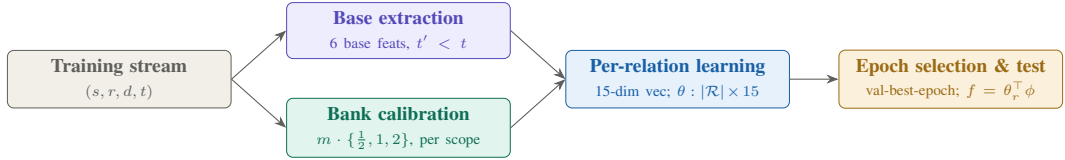

EdgeReMIND scores each candidate destination with a linear function of memorization features, using one weight vector per relation. It extracts six base count and recency features at three scopes (\emph{base extraction}), augments the recency signal with a calibrated multi-timescale bank (\emph{bank calibration}) to form a fixed 15-dimensional vector, and learns the per-relation weights by gradient descent (\emph{per-relation learning}). Figure~\ref{fig:pipeline} summarizes these three phases, and the subsections below detail each in turn. Appendix~\ref{app:components} expands the calibration and per-relation-weight stages with full per-dataset quantities, and Appendix~\ref{app:algorithms} gives the complete pseudocode: base extraction (Algorithm~\ref{alg:features}), bank calibration (Algorithm~\ref{alg:calibrate}), and per-relation learning (Algorithm~\ref{alg:train}).

\subsection{Problem Setup and Streaming Protocol}
\label{sec:problem}

A temporal multi-relational graph is a stream of timestamped quadruples $(s, r, d, t)$ with source $s$, relation $r \in \mathcal{R}$, destination $d$, and time $t$; this is the training stream of Figure~\ref{fig:pipeline}. The two dataset families differ in temporal resolution. The TKGs are discrete-time temporal graphs, whose facts are stamped at coarse day- to year-scale timesteps. The THGs are continuous-time temporal graphs, whose interactions carry fine seconds-scale timestamps and arrive as an event stream. Given a query $(s, r, ?, t)$, the task is to rank the true destination above a set of negative destinations. This work follows the TGB~2.0 streaming protocol~\citep{gastinger2024tgb2}: the prediction for a query at time $t$ depends only on edges with time $t' < t$. \S\ref{sec:impl} describes how both of EdgeReMIND's code paths enforce it.

\subsection{Base Memorization Features}
\label{sec:features}

Let $\mathcal{H}_t$ denote the multiset of edges observed before $t$. For a candidate destination $c$ under query $(s, r, t)$, six base features are computed (the base extraction stage of Figure~\ref{fig:pipeline}; Table~\ref{tab:features} summarizes all $15$ dimensions once the bank is defined). Throughout, a dot ($\cdot$) in a scope pattern is a wildcard that matches any value in that position. Three are unbounded counts:
\begin{align}
\phi_1 = \mathrm{cnt}_{srd} &= \big|\{(s,r,c)\in\mathcal{H}_t\}\big|,
\quad \phi_2 = \mathrm{cnt}_{rd} = \big|\{(\cdot,r,c)\in\mathcal{H}_t\}\big|,
\nonumber \\
\phi_3 = \mathrm{cnt}_{sd} &= \big|\{(s,\cdot,c)\in\mathcal{H}_t\}\big|.
\end{align}
Three are bounded recency signals in $[0, 1]$, computed from the most recent matching event preceding $t$:
\begin{align}
\phi_4 = \mathrm{dec}_{srd} &= \mathbf{1}_{(s,r,c) \in \mathcal{H}_t} \cdot e^{-\lambda (t - t^{\star}_{srd})},
\quad \phi_5 = \mathrm{dec}_{rd} = \mathbf{1}_{(\cdot,r,c) \in \mathcal{H}_t} \cdot e^{-\lambda (t - t^{\star}_{rd})},
\nonumber \\
\phi_6 = \mathrm{dec}_{d} &= \mathbf{1}_{(\cdot,\cdot,c) \in \mathcal{H}_t} \cdot e^{-\lambda_d (t - t^{\star}_{d})},
\end{align}
where $t^{\star}_X$ is the time of the most recent edge in $\mathcal{H}_t$ matching pattern $X$, $\lambda$ a fixed reference decay rate (set to a one-week rate for all datasets; see Table~\ref{tab:hparams} in Appendix~\ref{app:hparams}), $\lambda_d$ its destination-decay counterpart (equal to $\lambda$ by default), and $\mathbf{1}_{X}$ the indicator that any matching edge exists (else the feature is $0$). Intuitively, $\phi_1$ is exact source--relation--destination recurrence, $\phi_2$ is relation-conditioned destination popularity, $\phi_3$ is relation-agnostic source--destination affinity (the pairwise recurrence EdgeBank thresholds), and $\phi_4,\phi_5,\phi_6$ are bounded recency signals at progressively broader scopes: triple, relation--destination, and destination. The third recency feature $\phi_6$ is a bounded reformulation of PopTrack-style~\citep{gastinger2024tgb2} global destination popularity. These recency features are bounded to $[0, 1]$ by design: each depends only on the time since the key was last observed, giving the linear layer a stable, comparable signal across relations and datasets. This bounding is what lets a single configuration transfer across datasets: raw recency magnitudes vary widely across the suite, but once normalized to $[0, 1]$ the same learned weights apply unchanged, so no per-dataset tuning is required.

\subsection{Multi-Timescale Recency Bank}
\label{sec:bank}

The bounded recency features $\phi_4, \phi_5, \phi_6$ all share a single decay rate $\lambda$ chosen in absolute time units. This single rate is the main limitation of the base method, because the natural recurrence timescale is both relation-dependent and dataset-dependent. The recency bank addresses this by augmenting each scope with three additional bounded recency features at distinct half-lives, calibrated from the data (the bank calibration stage of Figure~\ref{fig:pipeline}).

\paragraph{Per-scope inter-recurrence gap calibration.} For each scope $X \in \{srd, rd, d\}$, let $G_X$ be the multiset of positive time gaps between consecutive same-key occurrences on the training stream:
\begin{equation}
G_X = \{\, t_{i+1} - t_i \;:\; \text{events } i, i{+}1 \text{ share the same } X\text{-key},\; t_{i+1} > t_i \,\}.
\end{equation}
Three half-lives per scope are placed geometrically around the median gap $m_X = \operatorname{median}(G_X)$:
\begin{equation}
h^{\mathrm{geo}}_{X,j} = m_X \cdot \gamma^{\,c_j}, \qquad c_j = j - \tfrac{n+1}{2}, \quad j = 1,\dots,n,
\label{eq:geometric-cal}
\end{equation}
where $\gamma$ is the geometric spacing factor and the centered exponents $c_j$ are symmetric about zero so the middle column anchors exactly on the median. The deployed configuration uses $n=3$ and $\gamma=2$ (sensitivity analysis in Appendix~\ref{app:ablations}), giving the spacing $h^{\mathrm{geo}}_X = \{m_X/2,\, m_X,\, 2\,m_X\}$. The corresponding decay rates are $\lambda_{X,j} = \ln 2 / h_{X,j}$, and the bank adds nine bounded features:
\begin{equation}
\phi^{\mathrm{bank}}_{X,j} = \mathbf{1}_{X\text{-key}(c)\,\in\,\mathcal{H}_t} \cdot e^{-\lambda_{X,j} (t - t^{\star}_X)}, \qquad X \in \{srd, rd, d\}, \; j \in \{1,2,3\}.
\end{equation}
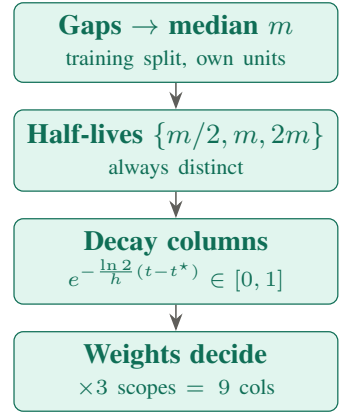
\begin{wrapfigure}{r}{0.36\linewidth}
\vspace{-8pt}
\centering
\resizebox{0.855\linewidth}{!}{%
\begin{tikzpicture}[
  node distance=3mm,
  cbox/.style={erbox, text width=36mm, minimum height=9mm, font=\footnotesize, inner sep=3pt, align=center},
]
\node[cbox, teal] (gap)
  {{\bfseries Gaps $\to$ median $m$}\\{\scriptsize training split, own units}};
\node[cbox, teal, below=of gap] (geo)
  {{\bfseries Half-lives $\{m/2,m,2m\}$}\\{\scriptsize always distinct}};
\node[cbox, teal, below=of geo] (dcols)
  {{\bfseries Decay columns}\\{\scriptsize $e^{-\frac{\ln 2}{h}(t-t^\star)}\!\in\![0,1]$}};
\node[cbox, teal, below=of dcols] (wt)
  {{\bfseries Weights decide}\\{\scriptsize $\times 3$ scopes $=9$ cols}};
\draw[erflow] (gap) -- (geo);
\draw[erflow] (geo) -- (dcols);
\draw[erflow] (dcols) -- (wt);
\end{tikzpicture}%
}
\caption{Bank calibration for one scope: three half-lives placed geometrically around the median gap, repeated over the three scopes to give the nine bank columns.}
\label{fig:calib}
\vspace{-10pt}
\end{wrapfigure}
Each bank feature reuses its base counterpart's last-seen timestamp $t^{\star}_X$, differing only in the decay rate.

\paragraph{Median anchoring and the spacing factor.} Anchoring on the median rather than a fixed absolute timescale is what makes the calibration unit-free. Because the median gap $m_X$ is in the dataset's own timestamp units, the spacing $\{m_X/2, m_X, 2m_X\}$ tracks the data's recurrence scale with no per-dataset constant, yielding minutes-to-days half-lives on seconds-scale event streams and year-scale half-lives on annually-timestamped graphs. The spacing factor $\gamma$ sets the width of this span. At $\gamma=1$ the three columns collapse to one; the deployed $\gamma=2$ separates them enough to capture distinct fast- and slow-recurring relations within a scope. This gives the per-relation learner independent short-, medium-, and long-timescale columns rather than a single rescaled decay. The spacing factor $\gamma$ and timescale count $n$ are the only hand-set hyperparameters, fixed across all eight datasets; calibration uses the training split only. The mechanism is depicted in Figure~\ref{fig:calib}; its per-dataset half-lives and empirical effect are reported in \S\ref{sec:calibration-empirics} and Appendix~\ref{app:calibration}.

\begin{table}[tbp]
\centering
\caption{The complete $15$-dimensional per-candidate feature vector $\phi(s,r,c,t)$, for query $(s,r,t)$ and candidate destination $c$. Scopes: $srd=(s,r,c)$, $rd=(\cdot,r,c)$, $sd=(s,\cdot,c)$, $d=(\cdot,\cdot,c)$; $t^{\star}_X$ is the last time scope $X$ matched, $\mathbf{1}_X$ the indicator that it ever did (feature is $0$ otherwise). Counts are unbounded; all recency and bank features lie in $[0,1]$. Base recency ($\phi_4$--$\phi_6$) uses a single fixed decay rate $\lambda$; each bank rate is $\lambda_{X,j}=\ln 2/h_{X,j}$ for a calibrated half-life $h_{X,j}\in\{m_X/2, m_X, 2m_X\}$ (\S\ref{sec:bank}).}
\label{tab:features}
\small
\setlength{\tabcolsep}{4pt}
\begin{tabular}{lllll}
\toprule
Index & Scope & Type & Definition & Meaning \\
\midrule
$\phi_1$ & $srd$ & count & $\big|\{(s,r,c)\in\mathcal{H}_t\}\big|$ & exact-triple recurrence \\
$\phi_2$ & $rd$ & count & $\big|\{(\cdot,r,c)\in\mathcal{H}_t\}\big|$ & relation--destination popularity \\
$\phi_3$ & $sd$ & count & $\big|\{(s,\cdot,c)\in\mathcal{H}_t\}\big|$ & source--destination affinity \\
$\phi_4$ & $srd$ & recency & $\mathbf{1}_{srd}\, e^{-\lambda (t - t^{\star}_{srd})}$ & exact-triple recency (fixed rate) \\
$\phi_5$ & $rd$ & recency & $\mathbf{1}_{rd}\, e^{-\lambda (t - t^{\star}_{rd})}$ & relation--destination recency (fixed rate) \\
$\phi_6$ & $d$ & recency & $\mathbf{1}_{d}\, e^{-\lambda_d (t - t^{\star}_{d})}$ & destination recency (fixed rate) \\
\midrule
$\phi^{\mathrm{bank}}_{srd,j}$ & $srd$ & bank & $\mathbf{1}_{srd}\, e^{-\lambda_{srd,j} (t - t^{\star}_{srd})}$ & exact-triple recency (three calibrated half-lives) \\
$\phi^{\mathrm{bank}}_{rd,j}$ & $rd$ & bank & $\mathbf{1}_{rd}\, e^{-\lambda_{rd,j} (t - t^{\star}_{rd})}$ & relation--destination recency (three calibrated half-lives) \\
$\phi^{\mathrm{bank}}_{d,j}$ & $d$ & bank & $\mathbf{1}_{d}\, e^{-\lambda_{d,j} (t - t^{\star}_{d})}$ & destination recency (three calibrated half-lives) \\
\bottomrule
\end{tabular}
\end{table}

\subsection{Per-Relation Learning}
\label{sec:learning}

A candidate is scored by a relation-specific linear combination of its features,
\begin{equation}
f(s, r, c, t) = \theta_r^{\top}\, \phi(s, r, c, t), \qquad \theta \in \mathbb{R}^{|\mathcal{R}| \times d},
\end{equation}
where $d = 6 + 9 = 15$ in the deployed configuration. The first six entries of $\theta_r$ (the base-feature weights) are initialized to $\omega_0 = (1.0, 10^{-3}, 10^{-2}, 2.0, 10^{-2}, 0.0)$, and the nine bank entries are initialized to $0$. The defaults are recurrence-plus-popularity weighted: the largest weights are on exact-triple recurrence ($\phi_1$) and its recency ($\phi_4$), with destination-popularity features present as tiebreakers. Because $\phi_6$ and the bank columns start at zero, any lift attributable to either comes purely from learning. An untrained model therefore produces exactly the same score as the bank-free baseline (up to floating-point summation order). Whether that learning is deployed well is the role of the selector (\S\ref{sec:selector}), with the per-dataset outcome in Figure~\ref{fig:contrib-base}.

The parameters $\theta$ are fit by minimizing, for each training query, a softmax cross-entropy over the true destination and $K=20$ relation-aware negative destinations, using Adam~\citep{kingma2015adam} at learning rate $\eta = 10^{-3}$ for $E = 30$ epochs. Because each relation owns its own weight vector across all 15 features, the model learns, per relation, both \emph{which signals matter} (recurrence, popularity, recency) and \emph{at which timescale the recency signal should be read} (short, medium, or long half-life). Training updates a $|\mathcal{R}| \times 15$ weight matrix, and the Adam optimization loop is a small fraction of total runtime, dominated instead by feature extraction.

\subsection{Validation-Best-Epoch Selection with Smoothing}
\label{sec:selector}

Choosing the training length $E$ per dataset would reintroduce the per-dataset tuning the method avoids. This is the epoch selection and test stage of Figure~\ref{fig:pipeline}. Instead, a snapshot of $\theta$ is captured after each epoch. After training, every snapshot is scored on the full validation set in a single multi-$\theta$ pass. This turns model selection into a byproduct of a single validation pass, a capability that follows directly from the features being independent of the learned weights. The memorization features depend only on $\mathcal{H}_t$, not on $\theta$, so one walk over the validation stream evaluates every snapshot at once. An embedding model cannot do this: it must re-run its network for each candidate. Selection uses the validation split only; the chosen snapshot is evaluated once on the test split, and every feature obeys the $t' < t$ streaming constraint of \S\ref{sec:problem}. The deployed model is the snapshot with the highest \emph{smoothed} validation MRR, a centered moving average over a window of $w$ epochs (default $w = 3$). The smoothing reduces sensitivity to single-epoch validation noise, which matters most where validation MRR is a less reliable predictor of test performance, such as the year-scale \texttt{tkgl-wikidata} (\S\ref{sec:wikidata-variance}). Selection therefore stays within the method's no-tuning discipline: it picks among learned snapshots on validation data alone, adding no per-dataset hyperparameter.

\subsection{Implementation and Scalability}
\label{sec:impl}
\label{sec:scalability}

EdgeReMIND runs on an \textbf{offline index} built on the Temporal Graph Modelling (TGM) library~\citep{chmura2025tgm}. The index precomputes the chronological edge stream once and stores each pattern scope ($(s,r,d)$, $(r,d)$, $(s,d)$, $d$) as a time-sorted array. Precomputing a global index is admissible because the streaming protocol is an information constraint, not a constraint on execution order: any implementation is valid so long as no edge at $t' \ge t$ influences the score. This conforms to the TGB streaming setting~\citep{huang2023tgb,gastinger2024tgb2}, which permits a model's memory to incorporate observed edges, including earlier edges of the test split, provided no weights are updated on test information. EdgeReMIND's memorization features are exactly such a memory: like EdgeBank~\citep{poursafaei2022edgebank}, they read past edges, and the learned weights $\theta_r$ are frozen after training and never updated at test time. A query at time $t$ retrieves only the prefix with $t' < t$ by binary search, so no edge with $t' \ge t$ can enter any feature value, in any partition. This single index serves training, validation, and test, and runs every experiment reported below. This guarantee is enforced by construction through the prefix lookup, and a separate streaming implementation corroborates it bit-for-bit on every feature column (Appendix~\ref{app:crosscheck}).

The offline-index design is trivially parallel: queries are mutually independent and dispatch to workers with no synchronization, and the read-only index is shared across workers with low memory overhead (Appendix~\ref{app:parallel}). Per-query scoring is a constant-time $15$-dimensional dot product. Each bank column reuses the last-seen time already computed for its base column, so the bank costs only a little extra arithmetic per query, not another index lookup. Appendix~\ref{app:parallel} gives the full design, and Appendix~\ref{app:code} covers implementation and code availability.

\section{Experiments}

\subsection{Setup}
\label{sec:setup}

Evaluation covers all eight datasets of TGB~2.0~\citep{gastinger2024tgb2}, loaded through py-tgb~2.2.0: four TKGs (\texttt{tkgl-smallpedia}, \texttt{tkgl-polecat}, \texttt{tkgl-icews}, \texttt{tkgl-wikidata}) and four THGs (\texttt{thgl-software}, \texttt{thgl-forum}, \texttt{thgl-github}, \texttt{thgl-myket}); per-dataset statistics are given in Appendix~\ref{app:datasets}. Filtered MRR is reported on the validation and test splits. For each query, other destinations that are also true at the query timestamp are removed from the candidate ranking before the target's reciprocal rank is computed. A model is therefore not penalized for ranking a genuinely correct alternative above the target (the time-aware filtered setting of TGB~2.0).

\paragraph{Hyperparameters.} A single configuration applies to every dataset, with no per-dataset tuning; the full set is listed in Appendix~\ref{app:hparams}.

\paragraph{Negative sampling.} TGB~2.0 uses three evaluation regimes. The three smaller TKGs (\texttt{tkgl-smallpedia}, \texttt{tkgl-polecat}, \texttt{tkgl-icews}) are 1-vs-all. The four THGs are 1-vs-$q$ with node-type-aware sampling. And \texttt{tkgl-wikidata} is 1-vs-1k with edge-type-aware sampling: the benchmark ranks against $1{,}000$ pre-generated negatives, a far smaller pool than 1-vs-all, which is why wikidata evaluates faster than the day-scale TKGs despite its size. All negatives are pre-generated by the benchmark and shared across methods. Training uses a separate type-constrained sampler that draws $K=20$ relation-aware negatives per query from the per-relation destination pool~\citep{krompass2015typeconstrained}, with uniform-random fallback when a pool is smaller than $K{+}1$. Because the regimes rank against candidate pools of different sizes, MRR is not directly comparable across datasets; Table~\ref{tab:recurrence} in Appendix~\ref{sec:relation-value} gives the per-query pool sizes.

\paragraph{Compute.} All runs used 64 CPU cores and 128~GB, with no GPU. Seeds were scheduled across a mix of CPU generations. MRR has no seed-level randomness for a fixed configuration, so accuracy differences reflect the method rather than sampling noise, while wall-clock varies modestly with the assigned node. Per-node hardware is summarized in Appendix~\ref{app:hardware}.

\subsection{Main Results: Five-Seed Sweep}
\label{sec:main-results}

Table~\ref{tab:main} reports validation and test MRR across all eight datasets together with per-seed wall-clock time. Five-seed sweeps (seeds $1337$--$1341$) are reported as mean $\pm$ unbiased standard deviation.

\begin{table}[!ht]
\centering
\caption{Per-dataset validation and test MRR under the deployed bank calibration. Val MRR is measured on the validation split (used only for epoch selection, \S\ref{sec:selector}); Test MRR is the reported held-out result on the test split. The validation and test columns cover disjoint chronological windows (test scored with more accumulated history), so the two are not directly comparable to each other. Entries are mean $\pm$ unbiased standard deviation over five seeds ($1337$--$1341$). Wall-clock and peak memory are per-seed (averaged over seeds for memory) and vary modestly with the assigned CPU node; MRR is deterministic for a fixed configuration. Compute and per-seed hardware detail are in Appendix~\ref{app:hardware}. Across datasets, MRR is not comparable between the three negative-sampling regimes; per-query candidate-pool sizes are given in Table~\ref{tab:recurrence} in Appendix~\ref{sec:relation-value}.}
\label{tab:main}
\footnotesize
\setlength{\tabcolsep}{4pt}\renewcommand{\arraystretch}{0.95}
\resizebox{\linewidth}{!}{%
\begin{tabular}{llrrrr}
\toprule
Dataset & Neg.\ sampling & Wall-clock (per seed) & Peak mem & Val MRR & Test MRR \\
\midrule
\multicolumn{6}{l}{\textit{Temporal Heterogeneous Graphs}}\\
\texttt{thgl-software} & 1-vs-$q$, node-type & $\sim 2$\,min & $14.8$~GB & $0.4509 \pm 0.0005$ & $0.5013 \pm 0.0007$ \\
\texttt{thgl-forum}    & 1-vs-$q$, node-type & $21$--$41$\,min & $30.8$~GB & $0.7336 \pm 0.0018$ & $\mathbf{0.7306 \pm 0.0021}$ \\
\texttt{thgl-github}   & 1-vs-$q$, node-type & $22$--$23$\,min & $32.6$~GB & $0.6953 \pm 0.0008$ & $\mathbf{0.7842 \pm 0.0008}$ \\
\texttt{thgl-myket}    & 1-vs-$q$, node-type & $1.6$--$3.7$\,h & $64.7$~GB & $0.9061 \pm 0.0006$ & $\mathbf{0.9056 \pm 0.0004}$ \\
\midrule
\multicolumn{6}{l}{\textit{Temporal Knowledge Graphs}}\\
\texttt{tkgl-smallpedia} & 1-vs-all           & $\sim 14$\,min & $13.6$~GB & $0.6554 \pm 0.0009$ & $\mathbf{0.6136 \pm 0.0017}$ \\
\texttt{tkgl-polecat}    & 1-vs-all           & $1.1$--$2.4$\,h & $69.6$~GB & $0.1769 \pm 0.0049$ & $0.1707 \pm 0.0046$ \\
\texttt{tkgl-icews}      & 1-vs-all           & $6.8$--$13.1$\,h & $73.4$~GB & $0.2615 \pm 0.0059$ & $\mathbf{0.2694 \pm 0.0034}$ \\
\texttt{tkgl-wikidata}   & 1-vs-1k, edge-type   & $20$--$35$\,min & $63.0$~GB & $0.7146 \pm 0.0029$ & $\mathbf{0.6397 \pm 0.0082}$ \\
\bottomrule
\end{tabular}%
}
\end{table}

\subsection{Comparison with the TGB~2.0 Leaderboard}
\label{sec:leaderboard}

Table~\ref{tab:leaderboard} places EdgeReMIND against methods reported on the TGB~2.0 leaderboards under identical evaluation protocols. EdgeReMIND test entries are five-seed means (seeds $1337$--$1341$; Table~\ref{tab:main}). Competitor numbers are taken verbatim from the official leaderboard~\citep{gastinger2024tgb2}. EdgeReMIND is state-of-the-art on six of the eight datasets. On the three largest by node count (\texttt{thgl-github}, \texttt{thgl-myket}, \texttt{tkgl-wikidata}), EdgeReMIND runs where learned methods fail. The wins reflect real recurrence in the data, not an absence of competitors (Appendices~\ref{sec:relation-value} and~\ref{app:relation-ablation}). On \texttt{tkgl-icews} it wins by a clear margin over the running competitors ($0.269$ over the next-best $0.211$). On \texttt{tkgl-smallpedia} and \texttt{thgl-forum} it narrowly exceeds embedding methods that run on the dataset (over CEN and TGN edge-type, respectively). On the two datasets where it is not first, it stays competitive, ranked second on \texttt{thgl-software} behind STHN, and behind TLogic~\citep{liu2022tlogic} by $0.057$ MRR on \texttt{tkgl-polecat}.

\begin{table}[t]
\centering
\caption{Test MRR on the TGB~2.0 leaderboard, by dataset (heterogeneous-graph prefix \texttt{thgl-} and knowledge-graph prefix \texttt{tkgl-} omitted). Best per dataset in bold, second-best underlined. ``--- OOM ---'' and ``--- OOT ---'' mark methods that exceed the TGB~2.0 benchmark's resource budget on that dataset: the GPU memory of the evaluation hardware (\rev{up to 40~GB}) for OOM, and the benchmark's seven-day per-experiment time limit for OOT~\citep{gastinger2024tgb2}; ``--'' denotes no leaderboard entry for that graph family.}
\label{tab:leaderboard}
\footnotesize
\setlength{\tabcolsep}{4pt}\renewcommand{\arraystretch}{0.95}
\resizebox{\textwidth}{!}{%
\begin{tabular}{lcccccccc}
\toprule
& \multicolumn{4}{c}{\textit{Temporal Heterogeneous Graphs}} & \multicolumn{4}{c}{\textit{Temporal Knowledge Graphs}} \\
\cmidrule(lr){2-5}\cmidrule(lr){6-9}
Method & \texttt{software} & \texttt{forum} & \texttt{github} & \texttt{myket} & \texttt{smallpedia} & \texttt{polecat} & \texttt{icews} & \texttt{wikidata} \\
\midrule
STHN      & \textbf{0.731} & --- OOM ---    & --- OOM ---    & --- OOM ---    & --             & --             & --             & --             \\
TGN (edge type)        & 0.424          & \underline{0.729}          & --- OOM ---    & --- OOM ---    & --             & --             & --             & --             \\
TGN                    & 0.324          & 0.649          & --- OOM ---    & --- OOM ---    & --             & --             & --             & --             \\
CEN                    & --             & --             & --             & --             & \underline{0.612}          & 0.184          & 0.187          & --- OOM ---    \\
TLogic                 & --             & --             & --             & --             & 0.595          & \textbf{0.228} & 0.186          & --- OOM ---    \\
RE-GCN                 & --             & --             & --             & --             & 0.594          & 0.175          & 0.182          & --- OOM ---    \\
EdgeBank (unlimited)   & 0.449          & 0.617          & \underline{0.413}          & \underline{0.456}          & 0.333          & 0.045          & 0.009          & \underline{0.535}          \\
EdgeBank (tw)          & 0.288          & 0.534          & 0.374          & 0.245          & 0.353          & 0.056          & 0.020          & \underline{0.535}          \\
RecB (train)     & --             & --             & --             & --             & 0.605          & \underline{0.198}          & \underline{0.211}          & --- OOM ---    \\
RecB (default)   & 0.099          & 0.561          & --- OOT ---    & --- OOT ---    & 0.486          & 0.167          & 0.206          & --- OOM ---    \\
\midrule
EdgeReMIND (this work) & \underline{0.501}          & \textbf{0.731} & \textbf{0.784} & \textbf{0.906} & \textbf{0.614} & 0.171          & \textbf{0.269} & \textbf{0.640} \\
\bottomrule
\end{tabular}%
}
\end{table}

\subsection{Performance Decomposition}
\label{sec:learning-value}
\label{sec:calibration-empirics}
\label{sec:smallpedia}
This section decomposes where EdgeReMIND's predictive performance originates. The base memorization features supply a deterministic baseline; the multi-timescale bank and per-relation learning build upon it. Three views follow: the contribution of each feature group, how the bank's calibration trades off against learning across datasets, and how much training the weights adds over the untrained baseline.

\begin{figure}[t]
\centering
\begin{minipage}{0.56\textwidth}
\centering
\resizebox{\linewidth}{!}{%
\begin{tikzpicture}[x=9mm,y=1mm]
  \def\s{82} 
  \draw[erGray,line width=0.4pt] (-0.5,0) -- (7.7,0); 
  \draw[erGray,line width=0.4pt] (-0.5,{-0.06*\s}) -- (-0.5,{0.37*\s});
  \foreach \v in {0,0.05,0.1,0.15,0.2,0.25,0.3,0.35}{
    \draw[erGray,line width=0.3pt] (-0.55,{\v*\s}) -- (-0.5,{\v*\s});
    \node[erGray,font=\scriptsize,left=1pt] at (-0.55,{\v*\s}) {\v};
  }
  \node[erGray,font=\scriptsize,rotate=90,align=center] at (-1.75,{0.13*\s}) {$\Delta$ Test MRR when removed};
  \foreach \i/\c/\r/\b in {%
    0/0.3246/0.0228/0.0043,
    1/0.0076/0.0486/0.0262,
    2/0.0471/0.0924/0.0054,
    3/0.0258/0.0007/0.0000,
    4/0.0050/0.2188/0.1508,
    5/0.0021/0.0435/0.0011,
    6/0.0159/0.0226/0.0077,
    7/0.0279/0.0529/0.0341%
  }{
    \pgfmathsetmacro{\up}{0}
    \pgfmathsetmacro{\dn}{0}
    \ifdim\c pt>0pt
      \draw[fill=erCountFill,draw=erAmber,line width=0.3pt] (\i-0.34,{\up*\s}) rectangle (\i+0.34,{(\up+\c)*\s});
      \pgfmathsetmacro{\up}{\up+\c}
    \else
      \draw[fill=erCountFill,draw=erAmber,line width=0.3pt] (\i-0.34,{\dn*\s}) rectangle (\i+0.34,{(\dn+\c)*\s});
      \pgfmathsetmacro{\dn}{\dn+\c}
    \fi
    \ifdim\r pt>0pt
      \draw[fill=erRecencyFill,draw=erTeal,line width=0.3pt] (\i-0.34,{\up*\s}) rectangle (\i+0.34,{(\up+\r)*\s});
      \pgfmathsetmacro{\up}{\up+\r}
    \else
      \draw[fill=erRecencyFill,draw=erTeal,line width=0.3pt] (\i-0.34,{\dn*\s}) rectangle (\i+0.34,{(\dn+\r)*\s});
      \pgfmathsetmacro{\dn}{\dn+\r}
    \fi
    \ifdim\b pt>0pt
      \draw[fill=erBankFill,draw=erBankFill,line width=0.3pt] (\i-0.34,{\up*\s}) rectangle (\i+0.34,{(\up+\b)*\s});
      \pgfmathsetmacro{\up}{\up+\b}
    \else
      \draw[fill=erBankFill,draw=erBankFill,line width=0.3pt] (\i-0.34,{\dn*\s}) rectangle (\i+0.34,{(\dn+\b)*\s});
      \pgfmathsetmacro{\dn}{\dn+\b}
    \fi
  }
  \foreach \i/\name in {0/software,1/forum,2/github,3/myket,4/smallpedia,5/polecat,6/icews,7/wikidata}{
    \node[erGray,font=\scriptsize,rotate=40,anchor=east] at (\i,{-0.065*\s}) {\texttt{\name}};
  }
  \draw[erGray,dashed,line width=0.3pt] (3.5,{-0.06*\s}) -- (3.5,{0.37*\s});
  \node[erGray,font=\scriptsize\itshape] at (1.5,{0.395*\s}) {Temporal Heterogeneous};
  \node[erGray,font=\scriptsize\itshape] at (5.7,{0.395*\s}) {Temporal Knowledge};
  \draw[erGray,line width=0.3pt] (4.5,{0.27*\s}) rectangle (6.75,{0.37*\s});
  \draw[fill=erCountFill,draw=erAmber,line width=0.3pt] (4.6,{0.345*\s}) rectangle (4.78,{0.355*\s}); \node[erGray,font=\scriptsize,right=0.5mm] at (4.78,{0.35*\s}) {count $\phi_1$--$\phi_3$};
  \draw[fill=erRecencyFill,draw=erTeal,line width=0.3pt]  (4.6,{0.315*\s}) rectangle (4.78,{0.325*\s}); \node[erGray,font=\scriptsize,right=0.5mm] at (4.78,{0.32*\s}) {recency $\phi_4$--$\phi_6$};
  \draw[fill=erBankFill,draw=erBankFill,line width=0.3pt]  (4.6,{0.285*\s}) rectangle (4.78,{0.295*\s}); \node[erGray,font=\scriptsize,right=0.5mm] at (4.78,{0.29*\s}) {bank $\phi_7$--$\phi_{15}$};
\end{tikzpicture}%
}
\end{minipage}%
\hfill
\begin{minipage}{0.40\textwidth}
\centering
\scriptsize
\setlength{\tabcolsep}{3pt}
\begin{tabular}{lrrr}
\toprule
Dataset & $-$Count & $-$Recency & $-$Bank \\
\midrule
\texttt{software}   & $-0.3246$ & $-0.0228$ & $-0.0043$ \\
\texttt{forum}      & $-0.0076$ & $-0.0486$ & $-0.0262$ \\
\texttt{github}     & $-0.0471$ & $-0.0924$ & $-0.0054$ \\
\texttt{myket}      & $-0.0258$ & $-0.0007$ & $+0.0000$ \\
\midrule
\texttt{smallpedia} & $-0.0050$ & $-0.2188$ & $-0.1508$ \\
\texttt{polecat}    & $-0.0021$ & $-0.0435$ & $-0.0011$ \\
\texttt{icews}      & $-0.0159$ & $-0.0226$ & $-0.0077$ \\
\texttt{wikidata}   & $-0.0279$ & $-0.0529$ & $-0.0341$ \\
\bottomrule
\end{tabular}
\end{minipage}
\caption{Per-dataset group-ablation importance in the trained model. Each feature group, count ($\phi_1$--$\phi_3$), recency ($\phi_4$--$\phi_6$), and the bank ($\phi_7$--$\phi_{15}$), is removed as a whole and the model retrained; bar height is the resulting drop in test MRR. The companion table gives the signed $\Delta$ test MRR (negative: removing the group hurts). Dataset names are abbreviated throughout the figures (the \texttt{thgl-}/\texttt{tkgl-} family prefixes are omitted); full names appear in the text. Seed 1337.}
\label{fig:contrib-all}
\end{figure}
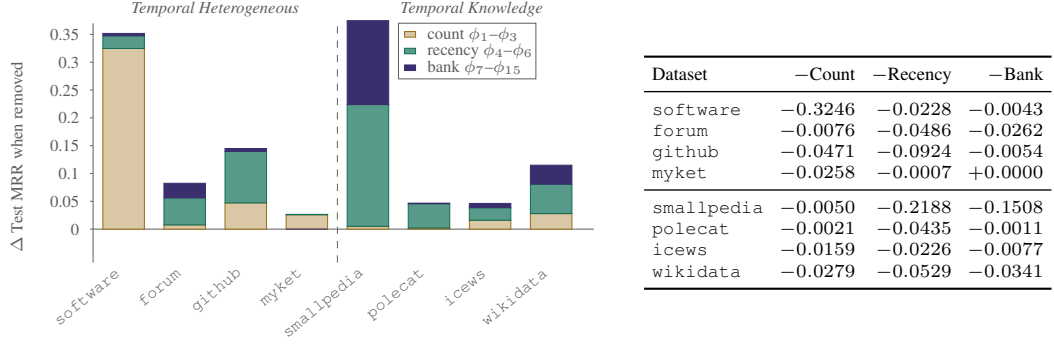

\paragraph{Group contributions, per dataset.} Figure~\ref{fig:contrib-all} shows how much each of the three groups is worth to the trained model, measured by ablating the whole group and retraining. Recency is the most broadly load-bearing group; counts carry \texttt{thgl-software} ($0.325$) and, to a lesser extent, \texttt{thgl-myket} ($0.026$); and the bank adds a substantial further increment on \texttt{tkgl-smallpedia} ($0.151$) and \texttt{tkgl-wikidata} ($0.034$), while contributing little elsewhere.

\paragraph{Bank calibration.}
The multi-timescale bank helps most on \texttt{tkgl-smallpedia}. That dataset's timestamps are in years and its inter-recurrence gap distribution is concentrated at exactly one year. At that scale the single fixed base decay rate collapses to a binary ``ever-seen'' indicator carrying no graded recency, so median-anchored geometric calibration replaces it with three bounded decay columns at $\{0.5, 1, 2\}$-year half-lives, letting the per-relation learner weight fast- and slow-recurring relations differently (full per-scope half-lives for all datasets in Appendix~\ref{app:calibration}). The bank-scope ablation (Appendix~\ref{app:ablations}) confirms the contribution is concentrated in the exact-triple (srd) scope, whose removal costs the largest single-scope decrement on this dataset.

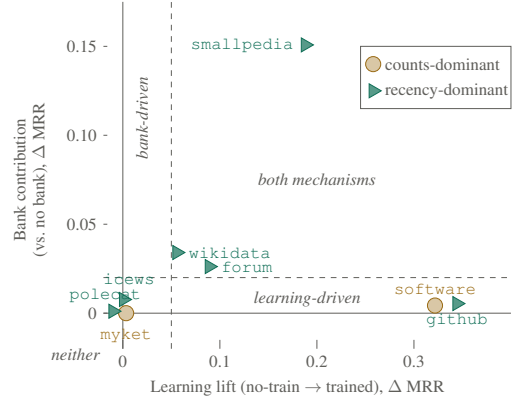
\begin{wrapfigure}{r}{0.50\textwidth}
\centering
\resizebox{0.48\textwidth}{!}{%
\begin{tikzpicture}[x=180mm,y=330mm]
  \def\xmin{-0.02}\def\xmax{0.40}\def\ymin{-0.02}\def\ymax{0.175}
  \draw[erGray,dashed,line width=0.3pt] (0.05,\ymin) -- (0.05,\ymax);
  \draw[erGray,dashed,line width=0.3pt] (\xmin,0.02) -- (\xmax,0.02);
  \draw[erGray,line width=0.4pt] (\xmin,0) -- (\xmax,0);
  \draw[erGray,line width=0.4pt] (0,\ymin) -- (0,\ymax);
  \foreach \v in {0,0.1,0.2,0.3}{
    \draw[erGray,line width=0.3pt] (\v,{\ymin}) -- (\v,{\ymin+0.005});
    \node[erGray,font=\small,below] at (\v,{\ymin}) {\v};
  }
  \foreach \v in {0,0.05,0.10,0.15}{
    \draw[erGray,line width=0.3pt] (\xmin,\v) -- (\xmin+0.004,\v);
    \node[erGray,font=\small,left] at (\xmin,\v) {\v};
  }
  \node[erGray,font=\small] at (0.18,{\ymin-0.022}) {Learning lift (no-train $\to$ trained), $\Delta$ MRR};
  \node[erGray,font=\small,rotate=90,align=center] at (\xmin-0.075,0.075) {Bank contribution\\(vs.\ no bank), $\Delta$ MRR};
  \node[erGray,font=\small\itshape,align=center] at (0.19,0.008) {learning-driven};
  \node[erGray,font=\small\itshape,align=center] at (0.20,0.075) {both mechanisms};
  \node[erGray,font=\small\itshape,rotate=90] at (0.02,0.11) {bank-driven};
  \node[erGray,font=\small\itshape,anchor=north east] at (-0.017,-0.016) {neither};
  \draw[fill=erCountFill,draw=erAmber,line width=0.3pt] (0.3217,0.0043) circle (1.4mm); 
  \draw[fill=erRecencyFill,draw=erTeal,line width=0.3pt] (0.0891,0.0262) +(0:1.5mm) -- +(120:1.5mm) -- +(240:1.5mm) -- cycle; 
  \draw[fill=erRecencyFill,draw=erTeal,line width=0.3pt] (0.3443,0.0054) +(0:1.5mm) -- +(120:1.5mm) -- +(240:1.5mm) -- cycle; 
  \draw[fill=erCountFill,draw=erAmber,line width=0.3pt] (0.0034,0.0000) circle (1.4mm);  
  \draw[fill=erRecencyFill,draw=erTeal,line width=0.3pt] (0.1881,0.1508) +(0:1.5mm) -- +(120:1.5mm) -- +(240:1.5mm) -- cycle; 
  \draw[fill=erRecencyFill,draw=erTeal,line width=0.3pt] (-0.0102,0.0011) +(0:1.5mm) -- +(120:1.5mm) -- +(240:1.5mm) -- cycle; 
  \draw[fill=erRecencyFill,draw=erTeal,line width=0.3pt] (0.0557,0.0341) +(0:1.5mm) -- +(120:1.5mm) -- +(240:1.5mm) -- cycle; 
  \draw[fill=erRecencyFill,draw=erTeal,line width=0.3pt] (0.0007,0.0077) +(0:1.5mm) -- +(120:1.5mm) -- +(240:1.5mm) -- cycle; 
  \node[erAmber,font=\small,above=0.8mm] at (0.3217,0.0043) {\texttt{software}};
  \node[erTeal,font=\small,right=1mm] at (0.0891,0.0262) {\texttt{forum}};
  \node[erTeal,font=\small,below=0.6mm] at (0.3443,0.0054) {\texttt{github}};
  \node[erAmber,font=\small,below=1.5mm] at (0.0034,0.0000) {\texttt{myket}};
  \node[erTeal,font=\small,left=1mm] at (0.1881,0.1508) {\texttt{smallpedia}};
  \node[erTeal,font=\small,above=0.6mm,xshift=-4pt] at (-0.0102,0.0011) {\texttt{polecat}};
  \node[erTeal,font=\small,right=1mm] at (0.0557,0.0341) {\texttt{wikidata}};
  \node[erTeal,font=\small,above=1.5mm,xshift=3pt] at (0.0007,0.0077) {\texttt{icews}};
  \draw[erGray,line width=0.3pt] (0.245,0.115) rectangle (0.40,0.150);
  \draw[fill=erCountFill,draw=erAmber,line width=0.3pt] (0.258,0.140) circle (1.4mm);
  \node[erGray,font=\small,right=1mm] at (0.258,0.140) {counts-dominant};
  \draw[fill=erRecencyFill,draw=erTeal,line width=0.3pt] (0.258,0.125) +(0:1.5mm) -- +(120:1.5mm) -- +(240:1.5mm) -- cycle;
  \node[erGray,font=\small,right=1mm] at (0.258,0.125) {recency-dominant};
\end{tikzpicture}%
}
\caption{Learning lift ($x$) vs.\ bank contribution ($y$, the drop in test MRR when the whole bank is removed) per dataset, colored and shaped by dominant feature group from the group ablation of Figure~\ref{fig:contrib-all} (counts vs.\ recency); differing axis scales. Seed 1337.}
\label{fig:contrib-feature}
\end{wrapfigure}
Figure~\ref{fig:contrib-feature} places this bank contribution against per-relation learning per dataset, colored and shaped by dominant group from the group ablation above (Figure~\ref{fig:contrib-all}); the two mechanisms are largely independent. \texttt{thgl-software} and \texttt{thgl-github} are learning-driven (large lift, negligible bank benefit); \texttt{tkgl-smallpedia}, \texttt{tkgl-wikidata}, and \texttt{thgl-forum} draw on both; and \texttt{thgl-myket}, \texttt{tkgl-polecat}, and \texttt{tkgl-icews} sit near the origin, where the baseline already nearly solves the task. No dataset is bank-driven without also being learning-driven: the bank helps only where per-relation learning is already the main lever, consistent with its role as a targeted refinement of the recency signal.

\FloatBarrier
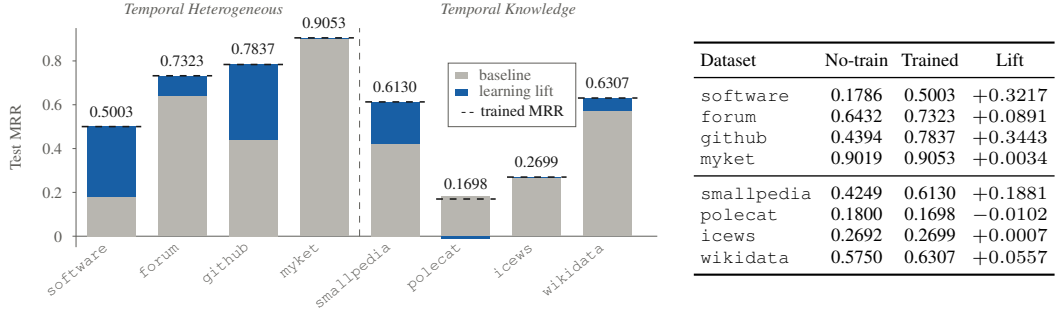
\begin{figure}[t]
\centering
\begin{minipage}{0.62\textwidth}
\centering
\resizebox{\linewidth}{!}{%
\begin{tikzpicture}[x=11mm,y=1mm]
  \def\s{34} 
  \draw[erGray,line width=0.4pt] (-0.5,0) -- (7.7,0);
  \draw[erGray,line width=0.4pt] (-0.5,{-0.05*\s}) -- (-0.5,{0.95*\s});
  \foreach \v in {0,0.2,0.4,0.6,0.8}{
    \draw[erGray,line width=0.3pt] (-0.55,{\v*\s}) -- (-0.5,{\v*\s});
    \node[erGray,font=\scriptsize,left=1pt] at (-0.55,{\v*\s}) {\v};
  }
  \node[erGray,font=\scriptsize,rotate=90] at (-1.35,{0.45*\s}) {Test MRR};
  \foreach \i/\base/\lift/\mrr in {%
    0/0.1786/0.3217/0.5003,
    1/0.6432/0.0891/0.7323,
    2/0.4394/0.3443/0.7837,
    3/0.9019/0.0034/0.9053,
    4/0.4249/0.1881/0.6130,
    5/0.1800/-0.0102/0.1698,
    6/0.2692/0.0007/0.2699,
    7/0.5750/0.0557/0.6307%
  }{
    \fill[erGrayL] (\i-0.34,0) rectangle (\i+0.34,{\base*\s});
    \ifdim\lift pt>0pt
      \fill[erBlue] (\i-0.34,{\base*\s}) rectangle (\i+0.34,{(\base+\lift)*\s});
    \else
      \fill[erBlue] (\i-0.34,0) rectangle (\i+0.34,{\lift*\s});
    \fi
    \draw[erInk,dashed,line width=0.7pt] (\i-0.42,{\mrr*\s}) -- (\i+0.42,{\mrr*\s});
    \node[erInk,font=\scriptsize,above] at (\i,{\mrr*\s+0.3}) {\mrr};
  }
  \foreach \i/\name in {0/software,1/forum,2/github,3/myket,4/smallpedia,5/polecat,6/icews,7/wikidata}{
    \node[erGray,font=\scriptsize,rotate=40,anchor=east] at (\i,{-0.015*\s}) {\texttt{\name}};
  }
  \draw[erGray,dashed,line width=0.3pt] (3.5,0) -- (3.5,{0.95*\s});
  \node[erGray,font=\scriptsize\itshape] at (1.3,{1.03*\s}) {Temporal Heterogeneous};
  \node[erGray,font=\scriptsize\itshape] at (5.6,{1.03*\s}) {Temporal Knowledge};
  \draw[erGray,line width=0.3pt] (4.75,{0.50*\s}) rectangle (6.55,{0.79*\s});
  \fill[erGrayL] (4.85,{0.72*\s}) rectangle (5.03,{0.75*\s}); \node[erGray,font=\scriptsize,right=0.5mm] at (5.03,{0.735*\s}) {baseline};
  \fill[erBlue] (4.85,{0.64*\s}) rectangle (5.03,{0.67*\s}); \node[erGray,font=\scriptsize,right=0.5mm] at (5.03,{0.655*\s}) {learning lift};
  \node[erInk,font=\scriptsize,anchor=west] at (4.82,{0.575*\s}) {-\,-\, trained MRR};
\end{tikzpicture}%
}
\end{minipage}%
\hfill
\begin{minipage}{0.35\textwidth}
\centering
\scriptsize
\setlength{\tabcolsep}{2.5pt}
\begin{tabular}{lccc}
\toprule
Dataset & No-train & Trained & Lift \\
\midrule
\texttt{software}   & 0.1786 & 0.5003 & $+0.3217$ \\
\texttt{forum}      & 0.6432 & 0.7323 & $+0.0891$ \\
\texttt{github}     & 0.4394 & 0.7837 & $+0.3443$ \\
\texttt{myket}      & 0.9019 & 0.9053 & $+0.0034$ \\
\midrule
\texttt{smallpedia} & 0.4249 & 0.6130 & $+0.1881$ \\
\texttt{polecat}    & 0.1800 & 0.1698 & $-0.0102$ \\
\texttt{icews}      & 0.2692 & 0.2699 & $+0.0007$ \\
\texttt{wikidata}   & 0.5750 & 0.6307 & $+0.0557$ \\
\bottomrule
\end{tabular}
\end{minipage}
\caption{Each dataset's test MRR split into the memorization baseline (gray) and the learning lift on top (blue; downward where learning lowers MRR). The dashed line marks the trained MRR; the companion table gives the exact no-train, trained, and lift values. Seed 1337.}
\label{fig:contrib-base}
\end{figure}

\paragraph{Per-relation learning.} The group and bank views ablate features; this view isolates the effect of training itself, comparing the learned weights against the untrained baseline (Figure~\ref{fig:contrib-base}). Learning adds the largest test-MRR gains on \texttt{thgl-github} ($+0.344$), \texttt{thgl-software} ($+0.322$), and \texttt{tkgl-smallpedia} ($+0.188$). Only the last of these also draws on the bank; the two THGs are learning-driven with negligible bank benefit (Figure~\ref{fig:contrib-feature}), confirming that the two mechanisms operate independently. Because the bank is zero-initialized, the untrained model scores identically to the bank-free baseline: the lifts measure learning over the deterministic recurrence-plus-popularity default. The split is starkly dataset-dependent (Figure~\ref{fig:contrib-base}): on the recurrence-saturated \texttt{thgl-myket} and \texttt{tkgl-icews} the baseline supplies essentially all of the performance ($0.902$ of $0.905$ and $0.269$ of $0.270$, respectively), where simple memorization already nearly solves the task. In contrast, \texttt{thgl-software}, \texttt{thgl-github}, and \texttt{tkgl-smallpedia} draw a large share of their MRR from learning. The slight negative on \texttt{tkgl-polecat} independently confirms the training-evaluation mismatch analyzed in \S\ref{sec:polecat-icews}. This decomposition explains how a dataset can show little in the marginal contribution views yet still post a high absolute score: its performance lives in the baseline.

\paragraph{Further ablations.}\label{sec:ablations} Appendix~\ref{app:ablations} reports four further ablation families. Feature leave-one-out and bank-scope drop isolate the contribution of individual features and timescale scopes, and the calibration-scheme comparison justifies the deployed geometric spacing against alternatives. \rev{A relation-awareness factorial varies the feature scopes and the weight structure independently, decomposing the relation's contribution.}

\section{Conclusion}

EdgeReMIND sets the highest reported test MRR on six of eight TGB~2.0 datasets while running entirely on CPU from a single $|\mathcal{R}| \times 15$ weight matrix with no per-dataset tuning. It is at once the strongest method on most of the benchmark and the only relation-aware method that runs on all of it. It scales to the three largest datasets, where every embedding method exhausts memory or the time budget. The gains are not merely memorization: training the per-relation weights lifts accuracy on both graph types, showing that learned relation-specific weighting, not the fixed heuristics alone, drives much of the advantage. This challenges the assumption that strong performance on these benchmarks requires representation learning: a linear model with per-relation learned weights over data-calibrated multi-timescale memorization features is both more accurate and far cheaper at scale.

\paragraph{Limitations.}
\label{sec:limitations}
Built on memorization features, EdgeReMIND cannot generalize to genuinely novel patterns. Its accuracy tracks the recurrence structure of the data, so on datasets with substantial novel-pattern content the memorization ceiling binds. The spacing factor $\gamma$ and timescale count $n$ remain hand-set constants ($\gamma=2$, $n=3$) rather than calibrated or learned. \rev{All reported results are on TGB~2.0. Whether the method transfers to other benchmarks is therefore not established here. The evidence rests on the diversity within this suite, spanning two graph families, timestamp units from seconds to years, and relation counts from $2$ to $1{,}192$.}

\paragraph{Future work.} The most immediate direction follows from the limitations: calibrating or learning the currently hand-set $\gamma$ and $n$. A further direction is moving the method from benchmark to deployment, where its low cost and retraining-free operation suit production settings. Online operation is already supported in principle, since the memorization state advances causally as edges stream in and tracks a live graph with no retraining; the weights could be re-fit incrementally on a sliding window, with decay rates recalibrated from the evolving gaps.

\makeatletter
\if@logreview\else
\subsubsection*{Acknowledgements}

The author sincerely thanks Dr.~Shenyang Huang and the TGB/TGM team for establishing the Temporal Graph Benchmark and its accompanying tools, and for the Temporal Graph Reading Group and conference presentations that together make it straightforward for researchers to contribute to this line of work.

Thanks also to the author's son, Bryant Pollard~II, whose accomplishments in machine learning research were the inspiration for this work, and whose review of this paper is deeply appreciated.

The author is grateful to Professor Brian Dean for his high-level review of this paper and for the algorithmic foundation his teaching provided.

This work was supported by resources provided by the Palmetto~2 cluster at Clemson University, with thanks to the Clemson University Research Computing and Data team for their support and for access to the Palmetto~2 high-performance computing infrastructure.

This research used in part resources on the Palmetto Cluster at Clemson University under National Science Foundation awards MRI 1228312, II NEW 1405767, MRI 1725573, and MRI 2018069~\citep{antao2024modernizing}. The views expressed in this article do not necessarily represent the views of NSF or the United States government.
\fi
\makeatother

\bibliographystyle{plainnat}
\bibliography{reference}

@inproceedings{kumar2019jodie,
  author    = {Kumar, Srijan and Zhang, Xikun and Leskovec, Jure},
  title     = {Predicting Dynamic Embedding Trajectory in Temporal Interaction Networks},
  booktitle = {Proceedings of the 25th ACM SIGKDD International Conference on Knowledge Discovery \& Data Mining (KDD)},
  pages     = {1269--1278},
  year      = {2019}
}

@inproceedings{poursafaei2022edgebank,
  author    = {Poursafaei, Farimah and Huang, Shenyang and Pelrine, Kellin and Rabbany, Reihaneh},
  title     = {Towards Better Evaluation for Dynamic Link Prediction},
  booktitle = {Advances in Neural Information Processing Systems (NeurIPS)},
  year      = {2022}
}

@inproceedings{huang2023tgb,
  author    = {Huang, Shenyang and Poursafaei, Farimah and Danovitch, Jacob and Fey, Matthias and Hu, Weihua and Rossi, Emanuele and Leskovec, Jure and Bronstein, Michael and Rabusseau, Guillaume and Rabbany, Reihaneh},
  title     = {Temporal Graph Benchmark for Machine Learning on Temporal Graphs},
  booktitle = {Advances in Neural Information Processing Systems (NeurIPS)},
  year      = {2023}
}

@inproceedings{gastinger2024tgb2,
  author    = {Gastinger, Julia and Huang, Shenyang and Galkin, Mikhail and Loghmani, Erfan and Parviz, Ali and Poursafaei, Farimah and Danovitch, Jacob and Rossi, Emanuele and Koutis, Ioannis and Stuckenschmidt, Heiner and Rabbany, Reihaneh and Rabusseau, Guillaume},
  title     = {{TGB 2.0}: A Benchmark for Learning on Temporal Knowledge Graphs and Heterogeneous Graphs},
  booktitle = {Advances in Neural Information Processing Systems (NeurIPS)},
  year      = {2024}
}

@inproceedings{gastinger2024recb,
  author    = {Gastinger, Julia and Meilicke, Christian and Errica, Federico and Sztyler, Timo and Schuelke, Anett and Stuckenschmidt, Heiner},
  title     = {History Repeats Itself: A Baseline for Temporal Knowledge Graph Forecasting},
  booktitle = {International Joint Conference on Artificial Intelligence (IJCAI)},
  year      = {2024},
  note      = {arXiv:2404.16726}
}

@misc{rossi2020tgn,
  author        = {Rossi, Emanuele and Chamberlain, Ben and Frasca, Fabrizio and Eynard, Davide and Monti, Federico and Bronstein, Michael},
  title         = {Temporal Graph Networks for Deep Learning on Dynamic Graphs},
  year          = {2020},
  eprint        = {2006.10637},
  archivePrefix = {arXiv},
  primaryClass  = {cs.LG},
  note          = {ICML Workshop on Graph Representation Learning}
}

@inproceedings{xu2020tgat,
  author    = {Xu, Da and Ruan, Chuanwei and Korpeoglu, Evren and Kumar, Sushant and Achan, Kannan},
  title     = {Inductive Representation Learning on Temporal Graphs},
  booktitle = {International Conference on Learning Representations (ICLR)},
  year      = {2020}
}

@misc{kazemi2019time2vec,
  author    = {Kazemi, Seyed Mehran and Goel, Rishab and Eghbali, Sepehr and Ramanan, Janahan and Sahota, Jaspreet and Thakur, Sanjay and Wu, Stella and Smyth, Cathal and Poupart, Pascal and Brubaker, Marcus},
  title     = {{Time2Vec}: Learning a Vector Representation of Time},
  year      = {2019},
  note      = {arXiv:1907.05321}
}

@inproceedings{kingma2015adam,
  author    = {Kingma, Diederik P. and Ba, Jimmy},
  title     = {{Adam}: A Method for Stochastic Optimization},
  booktitle = {International Conference on Learning Representations (ICLR)},
  year      = {2015}
}

@inproceedings{kondrup2025base3,
  author    = {Kondrup, Emma},
  title     = {{Base3}: A Simple Interpolation-Based Ensemble Method for Robust Dynamic Link Prediction},
  booktitle = {KDD Workshop on Temporal Graph Learning (TGL)},
  year      = {2025},
  note      = {arXiv:2506.12764}
}

@inproceedings{li2021regcn,
  author    = {Li, Zixuan and Jin, Xiaolong and Li, Wei and Guan, Saiping and Guo, Jiafeng and Shen, Huawei and Wang, Yuanzhuo and Cheng, Xueqi},
  title     = {Temporal Knowledge Graph Reasoning Based on Evolutional Representation Learning},
  booktitle = {International ACM SIGIR Conference on Research and Development in Information Retrieval (SIGIR)},
  year      = {2021}
}

@inproceedings{li2022cen,
  author    = {Li, Zixuan and Guan, Saiping and Jin, Xiaolong and Peng, Weihua and Lyu, Yajuan and Zhu, Yong and Bai, Long and Li, Wei and Guo, Jiafeng and Cheng, Xueqi},
  title     = {Complex Evolutional Pattern Learning for Temporal Knowledge Graph Reasoning},
  booktitle = {Annual Meeting of the Association for Computational Linguistics (ACL)},
  year      = {2022}
}

@inproceedings{liu2022tlogic,
  author    = {Liu, Yushan and Ma, Yunpu and Hildebrandt, Marcel and Joblin, Mitchell and Tresp, Volker},
  title     = {{TLogic}: Temporal Logical Rules for Explainable Link Forecasting on Temporal Knowledge Graphs},
  booktitle = {AAAI Conference on Artificial Intelligence (AAAI)},
  volume    = {36},
  pages     = {4120--4127},
  year      = {2022}
}

@inproceedings{li2023sthn,
  author    = {Li, Ce and Hong, Rongpei and Xu, Xovee and Trajcevski, Goce and Zhou, Fan},
  title     = {Simplifying Temporal Heterogeneous Network for Continuous-Time Link Prediction},
  booktitle = {ACM International Conference on Information and Knowledge Management (CIKM)},
  year      = {2023}
}

@inproceedings{krompass2015typeconstrained,
  author    = {Krompa{\ss}, Denis and Baier, Stephan and Tresp, Volker},
  title     = {Type-Constrained Representation Learning in Knowledge Graphs},
  booktitle = {The Semantic Web --- ISWC 2015},
  series    = {Lecture Notes in Computer Science},
  volume    = {9366},
  pages     = {640--655},
  publisher = {Springer},
  year      = {2015}
}

@misc{chmura2025tgm,
  author    = {Chmura, Jacob and Huang, Shenyang and Ngo, Tran Gia Bao and Parviz, Ali and Poursafaei, Farimah and Leskovec, Jure and Bronstein, Michael and Rabusseau, Guillaume and Fey, Matthias and Rabbany, Reihaneh},
  title     = {{TGM}: A Modular and Efficient Library for Machine Learning on Temporal Graphs},
  year      = {2025},
  note      = {arXiv:2510.07586}
}

@inproceedings{fan2021heterogeneous,
  title={Heterogeneous Temporal Graph Transformer: An Intelligent System for Evolving {A}ndroid Malware Detection},
  author={Fan, Yujie and Ju, Mingxuan and Hou, Shifu and Ye, Yanfang and Wan, Wenqiang and Wang, Kui and Mei, Yinming and Xiong, Qi},
  booktitle={Proceedings of the 27th ACM SIGKDD Conference on Knowledge Discovery \& Data Mining},
  pages={2831--2839},
  year={2021}
}

@inproceedings{song2020medicaltkg,
  title={Research of Medical Aided Diagnosis System Based on Temporal Knowledge Graph},
  author={Song, Feng and Wang, Bin and Tang, Yang and Sun, Jie},
  booktitle={International Conference on Advanced Data Mining and Applications (ADMA)},
  pages={236--250},
  year={2020},
  organization={Springer}
}

@inproceedings{antao2024modernizing,
  author    = {Antao, Asher and Burton, James Daly and Dawson, Douglas and Gemmill, Jill and Gerstener, Zachary and Godfrey, Ben and Groel, Scott and Jordan, Zach and Ligon, Becky and Smith, Dane and Moersen, Matthew and Nandigam, Sai and Rengier, Nicholas and Ryans, Dennis and Smith, Hudson and Wright, Thomas},
  title     = {Modernizing {C}lemson {U}niversity's {P}almetto Cluster: Lessons Learned from 17 Years of {HPC} Administration},
  booktitle = {Practice and Experience in Advanced Research Computing (PEARC '24)},
  articleno = {14},
  pages     = {1--9},
  publisher = {Association for Computing Machinery},
  year      = {2024},
  doi       = {10.1145/3626203.3670543}
}

\appendix
\renewcommand{\topfraction}{0.9}
\renewcommand{\bottomfraction}{0.7}
\renewcommand{\textfraction}{0.1}
\renewcommand{\floatpagefraction}{0.8}

\section{Reproducibility}
\label{app:repro}
This appendix documents everything needed to reproduce the reported results: the released code (\S\ref{app:code}), the full hyperparameter configuration (\S\ref{app:hparams}) with a single-run walkthrough (\S\ref{app:runinstructions}), and the hardware used (\S\ref{app:hardware}). All experiments use py-tgb~2.2.0.

\subsection{Code Availability}
\label{app:code}
EdgeReMIND is built on a copy of the TGM library~\citep{chmura2025tgm}, so that feature extraction, calibration, per-relation learning, and evaluation all run on the benchmark's common framework. It is not yet integrated into TGM but is intended for upstream contribution.
\makeatletter
\if@logreview
  An anonymized implementation is available at \url{https://anonymous.4open.science/r/EdgeReMIND-31B1}.
\else
  The implementation is available at \url{https://github.com/BryantPollard/EdgeReMIND}.
\fi
\makeatother

\subsection{Hyperparameter Configuration}
\label{app:hparams}

Table~\ref{tab:hparams} lists the full configuration. Every value is hand-set once and held identical across the benchmark.

\begin{table}[h]
\centering
\caption{Uniform hyperparameter configuration applied across all eight TGB~2.0 datasets. No per-dataset tuning.}
\label{tab:hparams}
\footnotesize
\setlength{\tabcolsep}{5pt}
\begin{tabular}{@{}l p{0.22\linewidth} p{0.34\linewidth}@{}}
\toprule
Hyperparameter & Value & Notes \\
\midrule
Training epochs $E$ & $30$ & val-best-epoch picks the deployed snapshot post-hoc \\
Optimizer & Adam~\citep{kingma2015adam} & \\
Learning rate $\eta$ & $10^{-3}$ & \\
Optimization batch size & $1024$ & \\
$L_2$ shrinkage on $\theta$ & $0$ & no prior toward deterministic defaults \\
Negatives per query $K$ & $20$ & relation-aware, in-pool~\citep{krompass2015typeconstrained} \\
Reference decay $\lambda$ ($\phi_4, \phi_5, \phi_6$) & $1/(7 \cdot 86400)$/sec & only used by base columns \\
Bank spacing factor $\gamma$, timescales $n$ & $\gamma=2$, $n=3$ & geometric, median-anchored; fixed default, not tuned (sensitivity in Table~\ref{tab:abl-phi}) \\
Default base weights $\omega_0$ & \multicolumn{2}{p{0.56\linewidth}}{$(1.0,\ 10^{-3},\ 10^{-2},\ 2.0,\ 10^{-2},\ 0.0)$, then zeros for the bank columns} \\
Smoothing window $w$ & $3$ & centered moving average on val MRR curve \\
Random seeds & $\{1337,\ldots,1341\}$ & fixed family, used uniformly \\
\bottomrule
\end{tabular}
\end{table}

\subsubsection{Reproducing a single run}
\label{app:runinstructions}

Every reported number comes from one of two example entry points, one per dataset family, invoked with the dataset name, seed, and (optionally) a worker count:
\begin{center}
\ttfamily python examples/linkproppred/(thgl|tkgl)/edgeremind.py --dataset <name> --seed <seed> [--num-workers <P>]
\end{center}
Here \texttt{(thgl|tkgl)} selects exactly one of the two dataset-family paths, \texttt{<...>} marks a required argument, and \texttt{[...]} an optional one. The \texttt{thgl} path serves the four THGs and the \texttt{tkgl} path the four TKGs, and \texttt{<seed>} ranges over $\{1337,\dots,1341\}$. Only the dataset and seed are required: every run uses the identical hand-set configuration of Table~\ref{tab:hparams} as the default, and datasets are downloaded automatically by the TGB benchmark on first invocation. The two quantities that differ by dataset are not passed as flags: the evaluation protocol is fixed by the TGB benchmark per dataset (recorded in Table~\ref{tab:main}), and the bank's decay half-lives are calibrated at runtime from each dataset's training gaps (Appendix~\ref{app:calibration}).

The worker-process parallelism that makes feature extraction scale (Appendix~\ref{app:parallel}) is exposed through \texttt{-{}-num-workers}~$P$. By default $P$ is auto-detected from the available core count, so results are correct with or without the flag; passing \texttt{-{}-num-workers}~$P$ explicitly is recommended to match the run to its environment (for example, the allocated cores on a cluster or a smaller worker count on a laptop), which is how the reported runs were launched. Reproducing the reported wall-clock times additionally requires the CPU-threading configuration used throughout, in which each numeric backend is pinned to a single thread and parallelism is taken at the worker-process level with $P$ the allocated core count. This single-thread cap is essential on multi-core clusters: without it, per-worker BLAS pools oversubscribe the cores and inflate runtime by more than an order of magnitude. The released code provides the full launch wrapper with its threading environment and worker configuration.

\subsection{Hardware}
\label{app:hardware}

All 40 main-results runs (Table~\ref{tab:main}) used a uniform scheduler allocation of 64 CPU cores and 128~GB of memory. The cluster allocator placed the large majority on Intel Xeon 8358 nodes, with a minority on AMD EPYC 9654/9655 and other Intel Xeon parts (8360Y, 8462Y+, 8470, 8580). EdgeReMIND is deterministic for a fixed configuration: repeated runs with the same seed, worker count, and software environment reproduce the reported MRR bit-for-bit, since the scores depend only on the memorization features and the learned per-relation weights. Test MRR is identical across the CPU generations above, which share that environment; only wall-clock and peak resident memory vary modestly with the assigned node, which is why those columns in Table~\ref{tab:main} are reported as ranges and seed-means. Small differences ($\lesssim 0.01$ MRR, and only on the most numerically-sensitive datasets) can arise across different software environments (PyTorch version, OS, or CPU/GPU build) because floating-point accumulation order changes; the reported numbers correspond to the documented environment (Linux, uv, PyTorch~2.5.1).

\section{Method Components in Detail}
\label{app:components}

This appendix details the three method stages: the base memorization-feature extraction for each query edge, the data-driven calibration that sets the recency bank's decay rates (\S\ref{sec:bank}), and the per-relation weight matrix the calibrated features feed into (\S\ref{sec:learning}), plus the parameter budget (\S\ref{app:parambudget}) and the parallelization design (\S\ref{app:parallel}). The complete $15$-dimensional feature vector is defined in Table~\ref{tab:features}.

\subsection{Algorithmic Summary}
\label{app:algorithms}

Algorithms~\ref{alg:features}--\ref{alg:train} give the full pseudocode for the three core phases of EdgeReMIND: $15$-dimensional feature construction, data-driven decay-rate calibration, and per-relation training, using the deployed configuration of Table~\ref{tab:hparams}.

\begin{algorithm}[htbp]
\small
\setstretch{0.95}
\caption{Base extraction: memorization features for one query edge.}
\label{alg:features}
\begin{algorithmic}[1]
\Require query $(s,r,d)$ at time $t$; history state $\mathcal{H}_t$ (per scope: last-seen time and prior-occurrence statistics); calibrated $\{\lambda^{(c)}_k\}$
\Ensure feature vector $x \in \mathbb{R}^{15}$; the $6$ base entries are $\phi_1,\dots,\phi_6$ of \S\ref{sec:features} (counts $\phi_1$--$\phi_3$ unbounded; recency $\phi_4$--$\phi_6$ and the $9$ bank entries in $[0,1]$)
\State \textbf{Base features (6):} $\phi_1,\dots,\phi_6$ as defined in \S\ref{sec:features}
  \Statex \hspace{1.5em} ($\phi_1$--$\phi_3$: nested-scope occurrence counts; $\phi_4$--$\phi_6$: bounded last-seen recency)
\For{each scope $c \in \{\mathrm{srd}, \mathrm{rd}, \mathrm{d}\}$} \Comment{recency bank: $3 \times n = 9$ features}
  \State $t_{\text{last}} \gets$ most recent time the scope key of $(s,r,d)$ appears in $\mathcal{H}_t$ (or $-\infty$)
  \State $\Delta t \gets t - t_{\text{last}}$
  \For{$k = 1$ to $n$}
    \State $x^{(c)}_k \gets \exp(-\lambda^{(c)}_k \,\Delta t)$ \Comment{bounded last-seen recency $\in[0,1]$; $0$ if never seen}
  \EndFor
\EndFor
\State $x \gets [\,\phi_1,\dots,\phi_6,\; x^{(\mathrm{srd})}_{1:n},\, x^{(\mathrm{rd})}_{1:n},\, x^{(\mathrm{d})}_{1:n}\,]$
\State \Return $x$
\end{algorithmic}
\end{algorithm}

\begin{algorithm}[htbp]
\small
\setstretch{0.95}
\caption{Bank calibration: per-scope decay rates from the training stream.}
\label{alg:calibrate}
\begin{algorithmic}[1]
\Require training edges $\{(s_i,r_i,d_i,t_i)\}_{i=1}^{N}$; timescale count $n$; spacing factor $\gamma$
\Ensure decay rates $\lambda^{(c)}_k$ for each scope $c\in\{\mathrm{srd},\mathrm{rd},\mathrm{d}\}$, $k=1{:}n$
\For{each scope $c \in \{\mathrm{srd}, \mathrm{rd}, \mathrm{d}\}$}
  \State group edges by the scope key ($(s,r,d)$ for srd, $(r,d)$ for rd, $(d)$ for d)
  \State $G_c \gets$ all positive inter-recurrence gaps $t_{j}-t_{j-1}$ within each group
  \State $m_c \gets \operatorname{median}(G_c)$ \Comment{data-driven anchor, in native time units}
  \For{$k = 1$ to $n$}
    \State $e_k \gets k - \tfrac{n+1}{2}$ \Comment{centered exponent}
    \State $h_{c,k} \gets m_c \cdot \gamma^{\,e_k}$ \Comment{geometric half-life spread around the median}
    \State $\lambda^{(c)}_k \gets \ln 2 \,/\, h_{c,k}$
  \EndFor
\EndFor
\State \Return $\{\lambda^{(c)}_k\}$
\end{algorithmic}
\end{algorithm}

\begin{algorithm}[htbp]
\small
\setstretch{0.95}
\caption{Per-relation learning: training the linear weights.}
\label{alg:train}
\begin{algorithmic}[1]
\Require training stream; validation stream; calibrated decay rates; epochs $E$; negatives per positive $K=20$
\Ensure per-relation weight matrix $\Theta \in \mathbb{R}^{|\mathcal{R}| \times 15}$ (row $\theta_r$ per relation)
\State calibrate $\{\lambda^{(c)}_k\}$ on the training stream \Comment{Alg.~\ref{alg:calibrate}}
\State build offline index $\mathcal{H}$ over train$+$val history
\State \textbf{initialize} every row $\theta_r \gets [\,\omega_0,\; \mathbf{0}_{9}\,]$, where
  \Statex \hspace{1.5em} $\omega_0 = (1.0,\,10^{-3},\,10^{-2},\,2.0,\,10^{-2},\,0.0)$ \Comment{base-feature default (\S\ref{sec:learning}); bank starts at zero}
\For{epoch $= 1$ to $E$}
  \For{each positive $(s,r,d,t)$ with relation-aware negatives $\{d^-_1,\dots,d^-_K\}$ \textbf{(in parallel)}}
    \State $x^+ \gets \textsc{Features}(s,r,d,t)$;\quad $x^-_j \gets \textsc{Features}(s,r,d^-_j,t)$ \Comment{Alg.~\ref{alg:features}}
    \State scores $z_0 \gets \theta_r^\top x^+$,\quad $z_j \gets \theta_r^\top x^-_j$ \Comment{linear scorer, no bias}
    \State $\mathcal{L} \mathrel{+}= -\log \dfrac{\exp(z_0)}{\exp(z_0) + \sum_{j=1}^{K}\exp(z_j)}$ \Comment{softmax cross-entropy}
  \EndFor
  \State update $\Theta$ by an Adam step on $\mathcal{L}$
\EndFor
\State select $\Theta$ at best smoothed validation MRR \Comment{\S\ref{sec:selector}}
\State \Return $\Theta$
\end{algorithmic}
\end{algorithm}

\FloatBarrier
\subsection{Full Calibration Detail}
\label{app:calibration}

The per-scope bank calibration (\S\ref{sec:calibration-empirics}) places the half-lives geometrically about each scope's median gap. The bank operates at three destination-containing scopes: exact-triple $(s,r,d)$, relation--destination $(r,d)$, and destination $(d)$. Each scope contains the destination because, under the ranking protocol, a feature discriminates between candidates only if its value changes with the candidate destination $c$. This three-scope selection is a design decision rather than an ablated one.

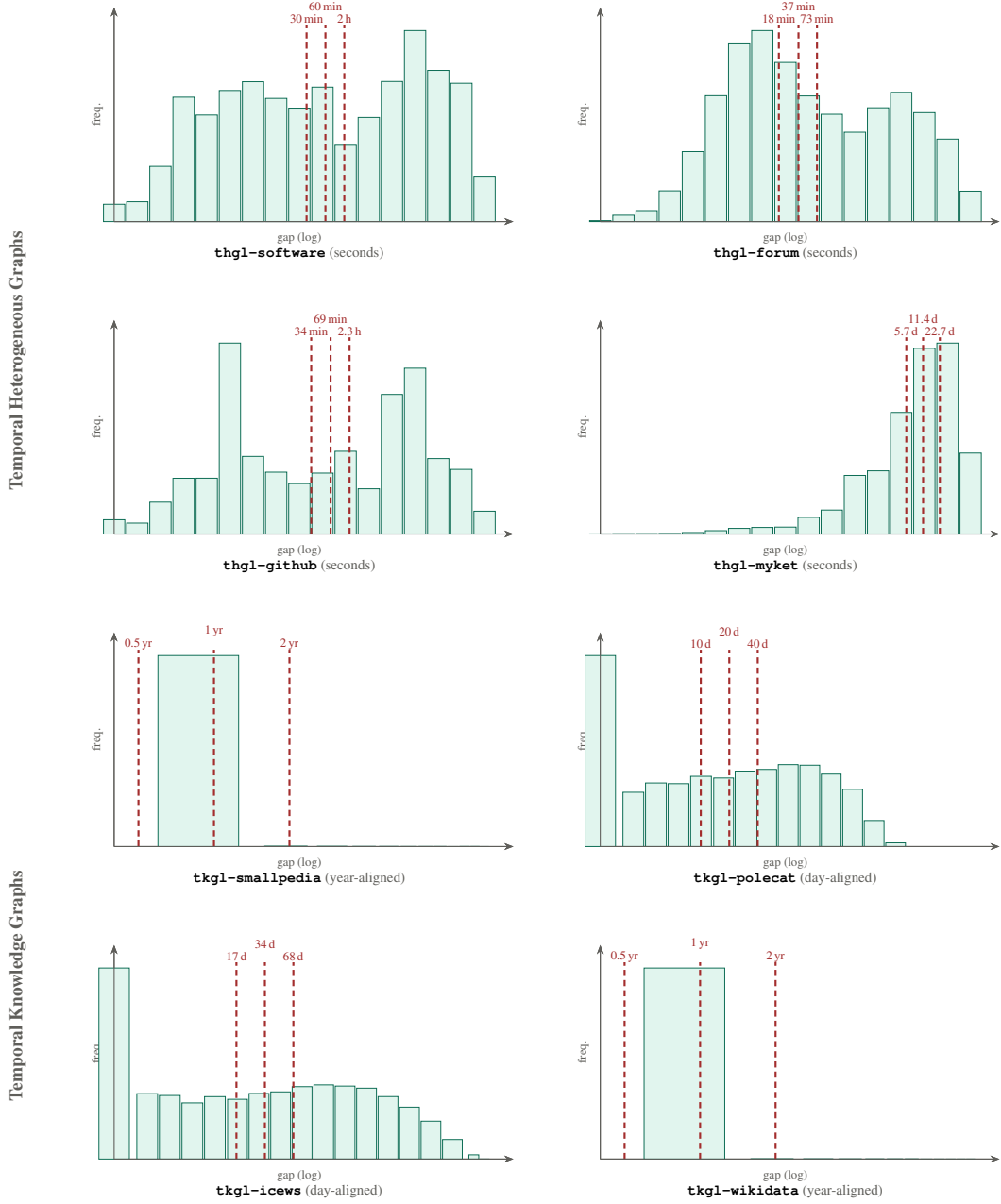
\begin{figure}[htbp]
\centering
\resizebox{\textwidth}{!}{%
\begin{tikzpicture}[font=\small,
  bar/.style={fill=erTealBg, draw=erTeal, line width=0.25pt},
  hl/.style={erRed, line width=1.0pt, dash pattern=on 3pt off 2pt},
  ax/.style={-{Stealth[length=1.6mm]}, draw=erGray, line width=0.5pt}]
\def\panelw{6.4}
\node[erGray,font=\bfseries\small,rotate=90,anchor=south] at (-1.4,13.800000000000004) {Temporal Heterogeneous Graphs};
\node[erGray,font=\bfseries\small,rotate=90,anchor=south] at (-1.4,3.0) {Temporal Knowledge Graphs};
\begin{scope}[xshift=0cm, yshift=16.200000000000003cm]
  \draw[ax] (0,0) -- ({\panelw+0.5},0);
  \node[below=2pt,font=\tiny,text=erGray] at ({\panelw/2},0) {gap (log)};
  \draw[ax] (0,0) -- (0,3.7);
  \node[left=2pt,font=\tiny,text=erGray,rotate=90,anchor=south] at (0,1.85) {freq.};
  \fill[bar] (-0.184,0) rectangle (0.184,0.299);
  \fill[bar] (0.216,0) rectangle (0.584,0.343);
  \fill[bar] (0.616,0) rectangle (0.984,0.954);
  \fill[bar] (1.016,0) rectangle (1.384,2.149);
  \fill[bar] (1.416,0) rectangle (1.784,1.840);
  \fill[bar] (1.816,0) rectangle (2.184,2.264);
  \fill[bar] (2.216,0) rectangle (2.584,2.416);
  \fill[bar] (2.616,0) rectangle (2.984,2.128);
  \fill[bar] (3.016,0) rectangle (3.384,1.959);
  \fill[bar] (3.416,0) rectangle (3.784,2.322);
  \fill[bar] (3.816,0) rectangle (4.184,1.318);
  \fill[bar] (4.216,0) rectangle (4.584,1.798);
  \fill[bar] (4.616,0) rectangle (4.984,2.420);
  \fill[bar] (5.016,0) rectangle (5.384,3.300);
  \fill[bar] (5.416,0) rectangle (5.784,2.612);
  \fill[bar] (5.816,0) rectangle (6.184,2.389);
  \fill[bar] (6.216,0) rectangle (6.584,0.783);
  \draw[hl] (3.325,0) -- (3.325,3.40);
  \node[erRed,font=\tiny] at (3.325,3.5) {30\,min};
  \draw[hl] (3.651,0) -- (3.651,3.40);
  \node[erRed,font=\tiny] at (3.651,3.72) {60\,min};
  \draw[hl] (3.977,0) -- (3.977,3.40);
  \node[erRed,font=\tiny] at (3.977,3.5) {2\,h};
  \node[font=\scriptsize\bfseries] at (3.2,-0.55) {\texttt{thgl-software}~{\scriptsize\mdseries\color{erGray}(seconds)}};
  \draw[ax] (0,0) -- ({\panelw+0.5},0);
  \draw[ax] (0,0) -- (0,3.7);
\end{scope}

\begin{scope}[xshift=8.4cm, yshift=16.200000000000003cm]
  \draw[ax] (0,0) -- ({\panelw+0.5},0);
  \node[below=2pt,font=\tiny,text=erGray] at ({\panelw/2},0) {gap (log)};
  \draw[ax] (0,0) -- (0,3.7);
  \node[left=2pt,font=\tiny,text=erGray,rotate=90,anchor=south] at (0,1.85) {freq.};
  \fill[bar] (-0.184,0) rectangle (0.184,0.016);
  \fill[bar] (0.216,0) rectangle (0.584,0.109);
  \fill[bar] (0.616,0) rectangle (0.984,0.189);
  \fill[bar] (1.016,0) rectangle (1.384,0.530);
  \fill[bar] (1.416,0) rectangle (1.784,1.209);
  \fill[bar] (1.816,0) rectangle (2.184,2.174);
  \fill[bar] (2.216,0) rectangle (2.584,3.070);
  \fill[bar] (2.616,0) rectangle (2.984,3.300);
  \fill[bar] (3.016,0) rectangle (3.384,2.749);
  \fill[bar] (3.416,0) rectangle (3.784,2.170);
  \fill[bar] (3.816,0) rectangle (4.184,1.853);
  \fill[bar] (4.216,0) rectangle (4.584,1.543);
  \fill[bar] (4.616,0) rectangle (4.984,1.965);
  \fill[bar] (5.016,0) rectangle (5.384,2.232);
  \fill[bar] (5.416,0) rectangle (5.784,1.881);
  \fill[bar] (5.816,0) rectangle (6.184,1.424);
  \fill[bar] (6.216,0) rectangle (6.584,0.524);
  \draw[hl] (3.086,0) -- (3.086,3.40);
  \node[erRed,font=\tiny] at (3.086,3.5) {18\,min};
  \draw[hl] (3.425,0) -- (3.425,3.40);
  \node[erRed,font=\tiny] at (3.425,3.72) {37\,min};
  \draw[hl] (3.745,0) -- (3.745,3.40);
  \node[erRed,font=\tiny] at (3.745,3.5) {73\,min};
  \node[font=\scriptsize\bfseries] at (3.2,-0.55) {\texttt{thgl-forum}~{\scriptsize\mdseries\color{erGray}(seconds)}};
  \draw[ax] (0,0) -- ({\panelw+0.5},0);
  \draw[ax] (0,0) -- (0,3.7);
\end{scope}

\begin{scope}[xshift=0cm, yshift=10.8cm]
  \draw[ax] (0,0) -- ({\panelw+0.5},0);
  \node[below=2pt,font=\tiny,text=erGray] at ({\panelw/2},0) {gap (log)};
  \draw[ax] (0,0) -- (0,3.7);
  \node[left=2pt,font=\tiny,text=erGray,rotate=90,anchor=south] at (0,1.85) {freq.};
  \fill[bar] (-0.184,0) rectangle (0.184,0.246);
  \fill[bar] (0.216,0) rectangle (0.584,0.188);
  \fill[bar] (0.616,0) rectangle (0.984,0.550);
  \fill[bar] (1.016,0) rectangle (1.384,0.963);
  \fill[bar] (1.416,0) rectangle (1.784,0.964);
  \fill[bar] (1.816,0) rectangle (2.184,3.300);
  \fill[bar] (2.216,0) rectangle (2.584,1.341);
  \fill[bar] (2.616,0) rectangle (2.984,1.069);
  \fill[bar] (3.016,0) rectangle (3.384,0.871);
  \fill[bar] (3.416,0) rectangle (3.784,1.053);
  \fill[bar] (3.816,0) rectangle (4.184,1.429);
  \fill[bar] (4.216,0) rectangle (4.584,0.782);
  \fill[bar] (4.616,0) rectangle (4.984,2.412);
  \fill[bar] (5.016,0) rectangle (5.384,2.866);
  \fill[bar] (5.416,0) rectangle (5.784,1.304);
  \fill[bar] (5.816,0) rectangle (6.184,1.116);
  \fill[bar] (6.216,0) rectangle (6.584,0.393);
  \draw[hl] (3.405,0) -- (3.405,3.40);
  \node[erRed,font=\tiny] at (3.405,3.5) {34\,min};
  \draw[hl] (3.740,0) -- (3.740,3.40);
  \node[erRed,font=\tiny] at (3.740,3.72) {69\,min};
  \draw[hl] (4.068,0) -- (4.068,3.40);
  \node[erRed,font=\tiny] at (4.068,3.5) {2.3\,h};
  \node[font=\scriptsize\bfseries] at (3.2,-0.55) {\texttt{thgl-github}~{\scriptsize\mdseries\color{erGray}(seconds)}};
  \draw[ax] (0,0) -- ({\panelw+0.5},0);
  \draw[ax] (0,0) -- (0,3.7);
\end{scope}

\begin{scope}[xshift=8.4cm, yshift=10.8cm]
  \draw[ax] (0,0) -- ({\panelw+0.5},0);
  \node[below=2pt,font=\tiny,text=erGray] at ({\panelw/2},0) {gap (log)};
  \draw[ax] (0,0) -- (0,3.7);
  \node[left=2pt,font=\tiny,text=erGray,rotate=90,anchor=south] at (0,1.85) {freq.};
  \fill[bar] (-0.184,0) rectangle (0.184,0.002);
  \fill[bar] (0.216,0) rectangle (0.584,0.005);
  \fill[bar] (0.616,0) rectangle (0.984,0.006);
  \fill[bar] (1.016,0) rectangle (1.384,0.009);
  \fill[bar] (1.416,0) rectangle (1.784,0.025);
  \fill[bar] (1.816,0) rectangle (2.184,0.057);
  \fill[bar] (2.216,0) rectangle (2.584,0.097);
  \fill[bar] (2.616,0) rectangle (2.984,0.113);
  \fill[bar] (3.016,0) rectangle (3.384,0.119);
  \fill[bar] (3.416,0) rectangle (3.784,0.287);
  \fill[bar] (3.816,0) rectangle (4.184,0.413);
  \fill[bar] (4.216,0) rectangle (4.584,1.012);
  \fill[bar] (4.616,0) rectangle (4.984,1.092);
  \fill[bar] (5.016,0) rectangle (5.384,2.101);
  \fill[bar] (5.416,0) rectangle (5.784,3.209);
  \fill[bar] (5.816,0) rectangle (6.184,3.300);
  \fill[bar] (6.216,0) rectangle (6.584,1.399);
  \draw[hl] (5.288,0) -- (5.288,3.40);
  \node[erRed,font=\tiny] at (5.288,3.5) {5.7\,d};
  \draw[hl] (5.578,0) -- (5.578,3.40);
  \node[erRed,font=\tiny] at (5.578,3.72) {11.4\,d};
  \draw[hl] (5.868,0) -- (5.868,3.40);
  \node[erRed,font=\tiny] at (5.868,3.5) {22.7\,d};
  \node[font=\scriptsize\bfseries] at (3.2,-0.55) {\texttt{thgl-myket}~{\scriptsize\mdseries\color{erGray}(seconds)}};
  \draw[ax] (0,0) -- ({\panelw+0.5},0);
  \draw[ax] (0,0) -- (0,3.7);
\end{scope}

\begin{scope}[xshift=0cm, yshift=5.4cm]
  \draw[ax] (0,0) -- ({\panelw+0.5},0);
  \node[below=2pt,font=\tiny,text=erGray] at ({\panelw/2},0) {gap (log)};
  \draw[ax] (0,0) -- (0,3.7);
  \node[left=2pt,font=\tiny,text=erGray,rotate=90,anchor=south] at (0,1.85) {freq.};
  \fill[bar] (0.757,0) rectangle (2.151,3.300);
  \fill[bar] (2.602,0) rectangle (3.335,0.006);
  \fill[bar] (3.511,0) rectangle (4.020,0.004);
  \fill[bar] (4.123,0) rectangle (4.515,0.002);
  \fill[bar] (4.585,0) rectangle (4.904,0.002);
  \fill[bar] (4.956,0) rectangle (5.225,0.001);
  \fill[bar] (5.266,0) rectangle (5.498,0.001);
  \fill[bar] (5.532,0) rectangle (5.737,0.001);
  \fill[bar] (5.766,0) rectangle (5.949,0.000);
  \fill[bar] (5.973,0) rectangle (6.139,0.001);
  \fill[bar] (6.160,0) rectangle (6.311,0.001);
  \fill[bar] (6.325,0) rectangle (6.475,0.000);
  \draw[hl] (0.420,0) -- (0.420,3.40);
  \node[erRed,font=\tiny] at (0.420,3.5) {0.5\,yr};
  \draw[hl] (1.725,0) -- (1.725,3.40);
  \node[erRed,font=\tiny] at (1.725,3.72) {1\,yr};
  \draw[hl] (3.029,0) -- (3.029,3.40);
  \node[erRed,font=\tiny] at (3.029,3.5) {2\,yr};
  \node[font=\scriptsize\bfseries] at (3.2,-0.55) {\texttt{tkgl-smallpedia}~{\scriptsize\mdseries\color{erGray}(year-aligned)}};
  \draw[ax] (0,0) -- ({\panelw+0.5},0);
  \draw[ax] (0,0) -- (0,3.7);
\end{scope}

\begin{scope}[xshift=8.4cm, yshift=5.4cm]
  \draw[ax] (0,0) -- ({\panelw+0.5},0);
  \node[below=2pt,font=\tiny,text=erGray] at ({\panelw/2},0) {gap (log)};
  \draw[ax] (0,0) -- (0,3.7);
  \node[left=2pt,font=\tiny,text=erGray,rotate=90,anchor=south] at (0,1.85) {freq.};
  \fill[bar] (-0.263,0) rectangle (0.263,3.300);
  \fill[bar] (0.392,0) rectangle (0.751,0.938);
  \fill[bar] (0.782,0) rectangle (1.141,1.099);
  \fill[bar] (1.172,0) rectangle (1.531,1.087);
  \fill[bar] (1.562,0) rectangle (1.920,1.214);
  \fill[bar] (1.960,0) rectangle (2.302,1.185);
  \fill[bar] (2.331,0) rectangle (2.675,1.304);
  \fill[bar] (2.706,0) rectangle (3.047,1.333);
  \fill[bar] (3.076,0) rectangle (3.419,1.416);
  \fill[bar] (3.451,0) rectangle (3.791,1.406);
  \fill[bar] (3.820,0) rectangle (4.161,1.252);
  \fill[bar] (4.190,0) rectangle (4.532,0.989);
  \fill[bar] (4.561,0) rectangle (4.903,0.450);
  \fill[bar] (4.933,0) rectangle (5.273,0.064);
  \draw[hl] (1.737,0) -- (1.737,3.40);
  \node[erRed,font=\tiny] at (1.737,3.5) {10\,d};
  \draw[hl] (2.229,0) -- (2.229,3.40);
  \node[erRed,font=\tiny] at (2.229,3.72) {20\,d};
  \draw[hl] (2.721,0) -- (2.721,3.40);
  \node[erRed,font=\tiny] at (2.721,3.5) {40\,d};
  \node[font=\scriptsize\bfseries] at (3.2,-0.55) {\texttt{tkgl-polecat}~{\scriptsize\mdseries\color{erGray}(day-aligned)}};
  \draw[ax] (0,0) -- ({\panelw+0.5},0);
  \draw[ax] (0,0) -- (0,3.7);
\end{scope}

\begin{scope}[xshift=0cm, yshift=0.0cm]
  \draw[ax] (0,0) -- ({\panelw+0.5},0);
  \node[below=2pt,font=\tiny,text=erGray] at ({\panelw/2},0) {gap (log)};
  \draw[ax] (0,0) -- (0,3.7);
  \node[left=2pt,font=\tiny,text=erGray,rotate=90,anchor=south] at (0,1.85) {freq.};
  \fill[bar] (-0.263,0) rectangle (0.263,3.300);
  \fill[bar] (0.392,0) rectangle (0.751,1.129);
  \fill[bar] (0.782,0) rectangle (1.141,1.098);
  \fill[bar] (1.172,0) rectangle (1.531,0.970);
  \fill[bar] (1.562,0) rectangle (1.920,1.078);
  \fill[bar] (1.960,0) rectangle (2.302,1.032);
  \fill[bar] (2.331,0) rectangle (2.675,1.132);
  \fill[bar] (2.706,0) rectangle (3.047,1.160);
  \fill[bar] (3.076,0) rectangle (3.419,1.250);
  \fill[bar] (3.451,0) rectangle (3.791,1.281);
  \fill[bar] (3.820,0) rectangle (4.161,1.260);
  \fill[bar] (4.190,0) rectangle (4.532,1.224);
  \fill[bar] (4.561,0) rectangle (4.903,1.079);
  \fill[bar] (4.933,0) rectangle (5.273,0.897);
  \fill[bar] (5.303,0) rectangle (5.644,0.654);
  \fill[bar] (5.674,0) rectangle (6.015,0.337);
  \fill[bar] (6.129,0) rectangle (6.300,0.073);
  \fill[bar] (6.315,0) rectangle (6.485,0.000);
  \draw[hl] (2.113,0) -- (2.113,3.40);
  \node[erRed,font=\tiny] at (2.113,3.5) {17\,d};
  \draw[hl] (2.605,0) -- (2.605,3.40);
  \node[erRed,font=\tiny] at (2.605,3.72) {34\,d};
  \draw[hl] (3.097,0) -- (3.097,3.40);
  \node[erRed,font=\tiny] at (3.097,3.5) {68\,d};
  \node[font=\scriptsize\bfseries] at (3.2,-0.55) {\texttt{tkgl-icews}~{\scriptsize\mdseries\color{erGray}(day-aligned)}};
  \draw[ax] (0,0) -- ({\panelw+0.5},0);
  \draw[ax] (0,0) -- (0,3.7);
\end{scope}

\begin{scope}[xshift=8.4cm, yshift=0.0cm]
  \draw[ax] (0,0) -- ({\panelw+0.5},0);
  \node[below=2pt,font=\tiny,text=erGray] at ({\panelw/2},0) {gap (log)};
  \draw[ax] (0,0) -- (0,3.7);
  \node[left=2pt,font=\tiny,text=erGray,rotate=90,anchor=south] at (0,1.85) {freq.};
  \fill[bar] (0.757,0) rectangle (2.151,3.300);
  \fill[bar] (2.602,0) rectangle (3.335,0.003);
  \fill[bar] (3.511,0) rectangle (4.020,0.002);
  \fill[bar] (4.123,0) rectangle (4.515,0.001);
  \fill[bar] (4.585,0) rectangle (4.904,0.001);
  \fill[bar] (4.956,0) rectangle (5.225,0.001);
  \fill[bar] (5.266,0) rectangle (5.498,0.001);
  \fill[bar] (5.532,0) rectangle (5.737,0.000);
  \fill[bar] (5.766,0) rectangle (5.949,0.000);
  \fill[bar] (5.973,0) rectangle (6.139,0.000);
  \fill[bar] (6.160,0) rectangle (6.311,0.000);
  \fill[bar] (6.325,0) rectangle (6.475,0.000);
  \draw[hl] (0.420,0) -- (0.420,3.40);
  \node[erRed,font=\tiny] at (0.420,3.5) {0.5\,yr};
  \draw[hl] (1.725,0) -- (1.725,3.40);
  \node[erRed,font=\tiny] at (1.725,3.72) {1\,yr};
  \draw[hl] (3.029,0) -- (3.029,3.40);
  \node[erRed,font=\tiny] at (3.029,3.5) {2\,yr};
  \node[font=\scriptsize\bfseries] at (3.2,-0.55) {\texttt{tkgl-wikidata}~{\scriptsize\mdseries\color{erGray}(year-aligned)}};
  \draw[ax] (0,0) -- ({\panelw+0.5},0);
  \draw[ax] (0,0) -- (0,3.7);
\end{scope}
\end{tikzpicture}%
}
\caption{Measured inter-recurrence gap distributions at the exact-triple (srd) scope only, and the geometric half-lives they yield, for all eight datasets in Table~\ref{tab:main} order (THGs above, TKGs below). Each panel is the histogram of training-stream srd-scope gaps $G_{srd}$ on a logarithmic gap axis, with the median-anchored half-lives $\{m/2, m, 2m\}$ (deployed $\gamma=2$) drawn as dashed lines at their calibrated values (Table~\ref{tab:halflives}). The other two scopes (rd, d) are not shown. Bins follow timestamp granularity: a per-panel logarithmic grid for the seconds-scale THGs, a shared day-aligned grid for \texttt{tkgl-polecat} and \texttt{tkgl-icews}, and a shared year-aligned grid for \texttt{tkgl-smallpedia} and \texttt{tkgl-wikidata} (only the populated head drawn). The year-scale graphs spike at the one-year gap; the others spread across orders of magnitude, several THGs multimodally (\texttt{thgl-github}, \texttt{thgl-forum} show a fast mode alongside a near-daily one). Time units on the dashed half-life labels: yr = years, d = days, h = hours, min = minutes.}
\label{fig:gapregime}
\end{figure}

Table~\ref{tab:halflives} reports, for the training split of every dataset's three scopes, the number of positive inter-recurrence gaps used for calibration and the resulting three half-lives. The single median-anchored geometric procedure adapts to each dataset's native time unit without per-dataset intervention, producing minute- to day-scale half-lives on the heterogeneous event streams, day-scale on \texttt{tkgl-icews} and \texttt{tkgl-polecat}, and year-scale on \texttt{tkgl-smallpedia} and \texttt{tkgl-wikidata}, and yields three half-lives per scope at $\{m/2, m, 2m\}$. Figure~\ref{fig:gapregime} shows this for the exact-triple scope per dataset. The concentrated year-scale graphs bunch the three half-lives near one value, while the spread distributions fan them out.

\begin{table}[h]
\centering
\caption{Deployed bank calibration (spacing factor $\gamma=2$) for all eight datasets: per-scope positive-gap count $|G_X|$, computed on the \emph{training} split only, and the resulting geometric half-lives $\{m_X/2, m_X, 2m_X\}$ in each dataset's own timestamp units. For the TKGs, the gaps are counted over the same inverse-inclusive stream as Table~\ref{tab:datastats}, so a scope's $|G_X|$ can exceed TGB's forward-only edge count. Time units: yr = years, d = days, h = hours, min = minutes.}
\label{tab:halflives}
\scriptsize
\setlength{\tabcolsep}{4pt}
\begin{tabular}{l l r l l l}
\toprule
Dataset & Scope & $|G_X|$ & $h_{X,1}=m_X/2$ & $h_{X,2}=m_X$ & $h_{X,3}=2m_X$ \\
\midrule
\multicolumn{6}{@{}l}{\textit{Temporal Heterogeneous Graphs}} \\
\texttt{thgl-software} & srd & $82{,}138$ & $30\,$min & $60\,$min & $2.0\,$h \\
 & rd & $403{,}064$ & $5.7\,$h & $11.5\,$h & $22.9\,$h \\
 & d & $682{,}753$ & $2.1\,$h & $4.2\,$h & $8.5\,$h \\
\midrule
\texttt{thgl-github} & srd & $126{,}837$ & $34\,$min & $69\,$min & $2.3\,$h \\
 & rd & $5{,}038{,}177$ & $77\,$min & $2.6\,$h & $5.1\,$h \\
 & d & $8{,}376{,}636$ & $8\,$min & $17\,$min & $33\,$min \\
\midrule
\texttt{thgl-forum} & srd & $9{,}568{,}325$ & $18\,$min & $37\,$min & $73\,$min \\
 & rd & $16{,}036{,}501$ & $2\,$min & $4\,$min & $8\,$min \\
 & d & $16{,}036{,}501$ & $2\,$min & $4\,$min & $8\,$min \\
\midrule
\texttt{thgl-myket} & srd & $8{,}438{,}605$ & $5.7\,$d & $11.4\,$d & $22.7\,$d \\
 & rd & $36{,}989{,}475$ & $3\,$min & $6\,$min & $13\,$min \\
 & d & $36{,}882{,}708$ & $2\,$min & $5\,$min & $10\,$min \\
\midrule
\midrule
\multicolumn{6}{@{}l}{\textit{Temporal Knowledge Graphs}} \\
\texttt{tkgl-smallpedia} & srd & $677{,}918$ & $0.5\,$yr & $1\,$yr & $2\,$yr \\
 & rd & $490{,}542$ & $0.5\,$yr & $1\,$yr & $2\,$yr \\
 & d & $440{,}127$ & $0.5\,$yr & $1\,$yr & $2\,$yr \\
\midrule
\texttt{tkgl-polecat} & srd & $910{,}314$ & $10\,$d & $20\,$d & $40\,$d \\
 & rd & $1{,}559{,}333$ & $4\,$d & $7\,$d & $14\,$d \\
 & d & $579{,}369$ & $2\,$d & $4\,$d & $8\,$d \\
\midrule
\texttt{tkgl-icews} & srd & $13{,}324{,}146$ & $17\,$d & $34\,$d & $68\,$d \\
 & rd & $15{,}663{,}712$ & $4\,$d & $8\,$d & $16\,$d \\
 & d & $7{,}175{,}332$ & $1\,$d & $2\,$d & $4\,$d \\
\midrule
\texttt{tkgl-wikidata} & srd & $11{,}962{,}006$ & $0.5\,$yr & $1\,$yr & $2\,$yr \\
 & rd & $8{,}531{,}287$ & $0.5\,$yr & $1\,$yr & $2\,$yr \\
 & d & $7{,}758{,}491$ & $0.5\,$yr & $1\,$yr & $2\,$yr \\
\bottomrule
\end{tabular}
\end{table}

\FloatBarrier

\subsection{Model Size and Parameter Budget}
\label{app:parambudget}

The deployed model learns a single $|\mathcal{R}| \times 15$ weight matrix $\theta$: one $15$-dimensional vector per relation type, over the six base and nine bank features. The model uses no node or edge embeddings and no other learned parameters, so the entire trainable footprint is exactly $15\,|\mathcal{R}|$. Table~\ref{tab:parambudget} lists $|\mathcal{R}|$ and the resulting parameter count for all eight datasets. The relation count spans nearly three orders of magnitude ($2$ to $1{,}192$), yet even the largest model has fewer than $18$k parameters. This is the concrete basis for the scalability claim of \S\ref{sec:scalability}: \texttt{tkgl-wikidata}, the $\sim$$10$M-fact graph on which every embedding-based method runs out of memory, is modeled here by a $17{,}880$-parameter linear layer that trains on CPU in seconds within a modest memory budget. The parameter budget grows only with the number of relation types, not with the number of nodes, edges, or timestamps, which is why the approach is insensitive to the graph scale that overwhelms representation-learning methods.

\begin{table}[h]
\centering
\caption{Relation count $|\mathcal{R}|$ and total learned parameters ($15\,|\mathcal{R}|$) per dataset. The model fits one $15$-dimensional weight vector per relation type and nothing else; the budget is independent of the number of nodes, edges, or timestamps. For the TKG datasets $|\mathcal{R}|$ is inverse-inclusive (Table~\ref{tab:datastats}).}
\label{tab:parambudget}
\footnotesize
\setlength{\tabcolsep}{8pt}
\begin{tabular}{lrr}
\toprule
Dataset & $|\mathcal{R}|$ & Learned params ($15\,|\mathcal{R}|$) \\
\midrule
\multicolumn{3}{@{}l}{\textit{Temporal Heterogeneous Graphs}} \\
\texttt{thgl-software}   & $14$   & $210$ \\
\texttt{thgl-forum}      & $2$    & $30$ \\
\texttt{thgl-github}     & $14$   & $210$ \\
\texttt{thgl-myket}      & $2$    & $30$ \\
\midrule
\multicolumn{3}{@{}l}{\textit{Temporal Knowledge Graphs}} \\
\texttt{tkgl-smallpedia} & $566$  & $8{,}490$ \\
\texttt{tkgl-polecat}    & $32$   & $480$ \\
\texttt{tkgl-icews}      & $782$  & $11{,}730$ \\
\texttt{tkgl-wikidata}   & $1{,}192$ & $17{,}880$ \\
\bottomrule
\end{tabular}
\end{table}

\subsection{Parallelization Design in Detail}
\label{app:parallel}

Feature extraction is embarrassingly parallel (\S\ref{sec:scalability}): a query's score depends only on its candidate features read against an immutable index, so a batch is partitioned across $W$ worker processes with no synchronization or shared mutable state. The index is built once as read-only NumPy arrays and shared across workers by a \texttt{ProcessPoolExecutor}. On Linux the pool uses the fork start method, so the read-only index is shared by copy-on-write and total memory stays at $O(|\text{index}|)$; on Windows and macOS the spawn start method is the default and each worker holds its own copy. All reported results were produced on Linux (fork). Feature extraction, validation snapshot scoring (one walk over the validation stream evaluates all 30 epoch snapshots at once), and test evaluation all parallelize this way; aside from the one-time index build, the only sequential cost is the TGB benchmark's negative-sampling hook, common to every method.

Per-query scoring is a constant-time $15$-dimensional dot product against the $|\mathcal{R}| \times 15$ matrix $\theta$ (at most tens of kilobytes). The bank adds negligible cost: each bank column reuses its base column's last-seen timestamp $t^{\star}_X$, so no extra binary searches are needed, only nine additional exponentials and a nine-element extension to the dot product. Calibration is a single pass over the training edges per scope and completes in seconds on the largest datasets.

\subsubsection{Streaming Cross-Check Against Leakage}
\label{app:crosscheck}
The offline index used for all experiments enforces streaming causality structurally: each query at time $t$ is answered from the $t' < t$ prefix of a time-sorted array, so an edge at $t' \ge t$ cannot enter any feature. EdgeReMIND also includes a second, independent feature implementation, used only for testing: a \textbf{streaming predictor} that maintains running memorization state incrementally in Python dictionaries, updating its state only after each query timestamp has passed. Because the two implementations share no code for feature computation, agreement between them is strong evidence that neither admits future information. The unit suite verifies that the index and the streaming predictor produce bit-identical feature values on every column, including all nine bank columns, across randomized synthetic streams and both zero and nonzero decay rates. Any leakage in the index would have to be exactly reproduced by the structurally different streaming path to escape this check.

\section{Datasets and Analysis}
\label{app:data}

This appendix collects the dataset-level material behind the results. Appendix~\ref{app:datasets} summarizes the eight datasets and their statistics. Appendix~\ref{sec:relation-value} is a cross-dataset analysis of how much conditioning on the relation adds over the destination alone. Appendix~\ref{app:perdataset} then examines how EdgeReMIND behaves on the individual datasets whose results most inform the method.

\subsection{Dataset Statistics}
\label{app:datasets}

Table~\ref{tab:datastats} summarizes each dataset. The datasets span roughly two orders of magnitude in edge count ($0.55$M to $53.6$M) and node count ($47$K to $5.9$M), with \texttt{thgl-github} carrying the most nodes and \texttt{thgl-myket} the most edges.

\begin{table}[ht]
\centering
\caption{Dataset scale statistics. Counts are over all splits; granularity is the median timestamp spacing and span the calendar range. TGB's TKG task adds an inverse relation per forward relation~\citep{gastinger2024tgb2}, so the stream scored doubles TGB's forward-only edge counts, and relation counts are the inverse-inclusive totals the model learns over (e.g.\ \texttt{tkgl-smallpedia}: $283$ forward, $566$ inverse-inclusive). This representation is TGB's own, so results are directly comparable to the leaderboard.}
\label{tab:datastats}
\footnotesize
\setlength{\tabcolsep}{5pt}
\begin{tabular}{lrrrrll}
\toprule
Dataset & \#Nodes & \#Edges & \#Rel. & \#Timestamps & Gran. & Span \\
\midrule
\multicolumn{7}{l}{\textit{Temporal Heterogeneous Graphs}} \\
\texttt{thgl-software}   & $681{,}927$   & $1{,}489{,}806$  & $14$    & $689{,}549$    & seconds & Jan 2024 \\
\texttt{thgl-forum}      & $152{,}816$   & $23{,}757{,}707$ & $2$     & $2{,}558{,}457$  & seconds & Jan 2014 \\
\texttt{thgl-github}     & $5{,}856{,}765$ & $17{,}499{,}577$ & $14$  & $2{,}510{,}415$  & seconds & Mar 2024 \\
\texttt{thgl-myket}      & $1{,}530{,}835$ & $53{,}632{,}788$ & $2$   & $14{,}828{,}090$ & seconds & Jun--Dec 2020 \\
\midrule
\multicolumn{7}{l}{\textit{Temporal Knowledge Graphs}} \\
\texttt{tkgl-smallpedia} & $47{,}433$    & $550{,}376$      & $566$   & $125$          & years   & 1900--2024 \\
\texttt{tkgl-polecat}    & $150{,}931$   & $1{,}779{,}610$  & $32$    & $1{,}826$      & days    & 2018--2022 \\
\texttt{tkgl-icews}      & $87{,}856$    & $15{,}513{,}446$ & $782$   & $10{,}224$     & days    & 1995--2022 \\
\texttt{tkgl-wikidata}   & $1{,}226{,}440$ & $9{,}856{,}203$ & $1{,}192$ & $2{,}025$    & years   & 0--2024 \\
\bottomrule
\end{tabular}
\end{table}

\subsection{The Value of the Relation}
\label{sec:relation-value}

The \emph{recurrence fraction} of a scope is the fraction of test edges whose key at that scope was already seen in the training-plus-validation history. It is the analogue of the exact-triple Recurrency Degree of TGB~2.0~\citep{gastinger2024tgb2}, extended to the model's four nested scopes: $(s,r,d)$, $(r,d)$, $(s,d)$, and $(d)$. Because it is measured over the model's inverse-included evaluation stream rather than the unique-fact test set, the exact-triple $(s,r,d)$ fraction runs lower than the Recurrency Degree reported by TGB~2.0 (for example, $0.267$ versus $0.61$ on \texttt{tkgl-wikidata}); the two track the same structure at different scales. Table~\ref{tab:recurrence} reports it for all eight datasets.

\begin{table}[!ht]
\centering
\caption{Recurrence fraction of test edges by scope: the fraction whose key was seen in the training-plus-validation history. Computed over the edge stream the model processes; for the TKG datasets this stream includes both directions of every fact, following TGB's standard representation (see Table~\ref{tab:datastats}), so $n_{\text{test}}$ is larger than the unique-fact test count. The last column isolates what conditioning on the relation adds over the destination alone: the retention $\mathrm{ret}=(r,d)/(d)$ is the fraction of destination recurrence retained when the relation must also match. Retention near $1$ means the relation is nearly redundant with the destination; see Appendix~\ref{sec:relation-value}.}
\label{tab:recurrence}
\footnotesize
\setlength{\tabcolsep}{5pt}
\begin{tabular}{lrrccccc}
\toprule
Dataset & $n_{\text{test}}$ & Neg./query & $(s,r,d)$ & $(r,d)$ & $(s,d)$ & $(d)$ & $\mathrm{ret}$ \\
\midrule
\multicolumn{8}{l}{\emph{1-vs-$q$ negative sampling}} \\
\texttt{thgl-software}   & $223{,}471$    & $1{,}000$ & $0.040$ & $0.383$ & $0.231$ & $0.561$ & $0.68$ \\
\texttt{thgl-forum}      & $3{,}563{,}653$ & $100$     & $0.515$ & $0.986$ & $0.515$ & $0.986$ & $1.00$ \\
\texttt{thgl-github}     & $2{,}624{,}932$ & $20$      & $0.006$ & $0.388$ & $0.082$ & $0.461$ & $0.84$ \\
\texttt{thgl-myket}      & $8{,}044{,}915$ & $20$      & $0.326$ & $0.991$ & $0.359$ & $0.992$ & $1.00$ \\
\midrule
\multicolumn{8}{l}{\emph{1-vs-all negative sampling}} \\
\texttt{tkgl-smallpedia} & $163{,}172$    & $47{,}433$  & $0.361$ & $0.789$ & $0.366$ & $0.827$ & $0.95$ \\
\texttt{tkgl-polecat}    & $532{,}636$    & $150{,}931$ & $0.338$ & $0.805$ & $0.501$ & $0.906$ & $0.89$ \\
\texttt{tkgl-icews}      & $4{,}651{,}378$ & $87{,}856$  & $0.371$ & $0.531$ & $0.812$ & $0.984$ & $0.54$ \\
\midrule
\multicolumn{8}{l}{\emph{1-vs-1k negative sampling}} \\
\texttt{tkgl-wikidata}   & $2{,}877{,}500$ & $1{,}000$ & $0.267$ & $0.614$ & $0.274$ & $0.652$ & $0.94$ \\
\bottomrule
\end{tabular}
\end{table} This section uses those fractions to ask a cross-dataset question about the data itself: does conditioning on the relation narrow where recurrence happens, or is the relation largely redundant with the destination? The relation-aware scope $(r,d)$ is a strict refinement of the destination-only scope $(d)$: seeing a $(r,d)$ pair implies seeing $d$, so $(r,d) \le (d)$ always. The retention $\mathrm{ret}=(r,d)/(d)$ is the fraction of destination recurrence that survives requiring the relation to match. Retention near $1$ means a destination that recurs almost always recurs under the same relation, so the relation is nearly redundant with the destination. Low retention means destinations recur under many relations.

On seven of the eight datasets retention is high ($0.84$ to $1.00$), and on the three largest by node count it is $0.84$ (\texttt{thgl-github}), $0.94$ (\texttt{tkgl-wikidata}), and $1.00$ (\texttt{thgl-myket}). Where retention is this high, a relation-agnostic method has almost the same recurrence to exploit as a relation-aware one. EdgeBank keys on the source--destination pair and popularity-based methods on the destination, and Base3~\citep{kondrup2025base3} fuses both, alongside a temporal co-occurrence memory. On \texttt{thgl-myket}, where destination recurrence is $99.2\%$ and retention is $1.00$, such a method would very likely score well without any relation term. The exception is \texttt{tkgl-icews} (retention $0.54$): its destinations recur under many different relations, so the relation carries genuine additional coverage there. \rev{Coverage is not the whole story. The relation-awareness factorial (\S\ref{app:relation-ablation}) shows that the relation's accuracy contribution does not track retention. It is largest on \texttt{thgl-github} and negligible on \texttt{tkgl-wikidata}, despite their similar retention. The factorial does bear out the coverage prediction. On \texttt{thgl-myket} and \texttt{thgl-forum}, both at retention $1.00$, the relation-blind model is within $0.005$ of the full model.}

The retention analysis is therefore a caution as much as a claim: on the high-retention datasets the relation adds little coverage beyond the destination, so any advantage over a scalable relation-agnostic baseline must come from precision on the minority of ambiguous cases, from the multi-timescale bank, or from learning (\S\ref{sec:learning-value}), rather than from raw recurrence.

\subsection{Per-Dataset Analysis}
\label{app:perdataset}

This appendix examines how EdgeReMIND behaves on the individual datasets whose results most inform the method, rather than the datasets in isolation. It covers \texttt{tkgl-wikidata}, which shows the suite's largest seed variance (\S\ref{sec:wikidata-variance}); and \texttt{tkgl-polecat} and \texttt{tkgl-icews}, where a training-evaluation distribution mismatch (not the calibration or features) bounds performance (\S\ref{sec:polecat-icews}).

\subsubsection{Wikidata: Stable Selection Under Year-Scale Drift}
\label{sec:wikidata-variance}

On \texttt{tkgl-wikidata}, validation MRR is a less trustworthy predictor of test MRR because the data drifts over its long, year-scale timeline, making it harder to reliably select the best model from validation. EdgeReMIND nonetheless reaches $0.6397 \pm 0.0082$ test MRR, the highest reported on this dataset. That standard deviation is the largest in the suite, but at under $1\%$ it is far below the $\sim$$0.10$ margin over the next method, so the ranking is secure.

\subsubsection{Polecat and ICEWS: Training-Evaluation Distribution Mismatch}
\label{sec:polecat-icews}

On \texttt{tkgl-polecat} and \texttt{tkgl-icews}, training the per-relation weights does not improve test MRR, and on \texttt{tkgl-polecat} it slightly reduces it relative to the deterministic configuration. The candidates seen during training differ from those ranked at evaluation. Relation-aware negative sampling draws $K=20$ hard, in-pool negatives that have co-occurred with the relation. But 1-vs-all evaluation ranks against the full destination set (the Neg./query column of Table~\ref{tab:recurrence}), most of which have never been seen by the model. Learning can then over-optimize the dense in-pool subproblem at marginal cost to the out-of-pool majority. The bank's zero-initialization is the safeguard.

Empirically the safeguard holds. Across the full five-seed sweep, \texttt{tkgl-polecat} reaches test MRR $0.1707 \pm 0.0046$, competitive with the memorization baselines on this dataset (Table~\ref{tab:leaderboard}). No seed regresses meaningfully: the worst ($0.1652$) and best ($0.1775$) bracket the reported mean. The stability is attributable to the val-best-epoch selector. It snapshots the weights before any such over-optimization sets in, so the trained model neither gains nor loses. \texttt{tkgl-icews} behaves similarly: five-seed test MRR $0.2694 \pm 0.0034$, the highest reported.

Both are day-scale TKGs whose gap distributions are not concentrated at a single value, so the three median-anchored half-lives span the natural range without the year-scale collapse. The low absolute MRR reflects a thin memorization signal, not a calibration or feature deficiency. Both datasets carry political-event content, in which facts rarely persist across consecutive steps: their Direct Recurrency Degree~\citep{gastinger2024tgb2} is the lowest in the suite (\texttt{tkgl-polecat} $0.07$, \texttt{tkgl-icews} $0.11$). This caps the memorization signal available, and the training--evaluation mismatch then governs how much of it the learner captures.

\rev{These two datasets differ in more than the training mismatch. In the factorial of \S\ref{app:relation-ablation}, collapsing the relation-conditioned feature scopes costs $0.101$ MRR on \texttt{tkgl-icews}, the second-largest scope effect in the table. That dataset has the lowest retention in the suite ($0.54$), and its destinations recur under many relations, so the relation carries coverage the destination alone does not. The corresponding entry for \texttt{tkgl-polecat} is one of the selector-unstable cells and is not compared here. Relation-conditioned features therefore benefit \texttt{tkgl-icews} even though training adds little there.}

\section{Additional Ablations}
\label{app:ablations}

This appendix supplements the no-train study of \S\ref{sec:learning-value} with four ablation families that remove parts of the model and remeasure: feature leave-one-out, bank-scope drop, calibration-scheme, and \rev{a relation-awareness factorial}. The first two are single-seed (1337) directional studies across all eight datasets, with deltas relative to the full-model seed-1337 run; because they are single-seed, small deltas ($|\Delta| \lesssim 0.01$) fall within the seed-to-seed and cross-environment variation observed in the five-seed runs, and only large, consistent effects should be read as structural. The calibration-scheme ablation and \rev{the relation-awareness factorial}, by contrast, are run at the full five seeds and reported as means (with standard deviation where relevant), so their deltas are interpreted against the per-seed standard deviation.

\paragraph{Feature leave-one-out: counts dominate THGs, recency dominates the year-scale TKGs.} Table~\ref{tab:abl-loo} drops the first six base features individually. The THGs are dominated by \emph{count} features: removing the source--destination count $\phi_3$ collapses \texttt{thgl-software} by $-0.299$, and \texttt{thgl-forum} and \texttt{thgl-myket} depend most on the relation--destination count $\phi_2$ (the one exception is \texttt{thgl-github}, which leans on the destination-recency feature $\phi_6$, $-0.055$). The year-scale TKGs are instead dominated by \emph{recency}: removing the triple-specific recency $\phi_4$ costs \texttt{tkgl-smallpedia} $-0.174$ and \texttt{tkgl-wikidata} $-0.029$, the largest drops on each, consistent with these datasets carrying graded fact-recurrence structure that a count alone cannot capture (the same structure the multi-timescale bank exploits, \S\ref{sec:smallpedia}). The day-scale TKGs are more mixed: \texttt{tkgl-icews} leans on the exact-triple count $\phi_1$ ($-0.047$) and \texttt{tkgl-polecat} on destination recency $\phi_6$ ($-0.014$), both with smaller magnitudes, reflecting their lower overall memorization ceiling. The broad pattern (heterogeneous interaction graphs driven by co-occurrence counts, year-scale TKGs by fact recency) is the clearest single-feature signal in the suite.

\begin{table}[!ht]
\centering
\caption{Feature leave-one-out (test MRR $\Delta$ vs.\ full model, seed 1337). Columns are the six base features; each entry drops that feature alone. The largest drop per row is in \textbf{bold}. $\phi_1$--$\phi_3$ are counts (srd, rd, sd); $\phi_4$--$\phi_6$ are bounded recency (srd, rd, d). Small positive values are single-seed noise.}
\label{tab:abl-loo}
\scriptsize
\setlength{\tabcolsep}{4pt}
\begin{tabular}{lcccccc}
\toprule
Dataset & $\phi_1$ cnt$_{srd}$ & $\phi_2$ cnt$_{rd}$ & $\phi_3$ cnt$_{sd}$ & $\phi_4$ dec$_{srd}$ & $\phi_5$ dec$_{rd}$ & $\phi_6$ dec$_{d}$ \\
\midrule
\multicolumn{7}{@{}l}{\textit{Temporal Heterogeneous Graphs}} \\
\texttt{thgl-software}   & $-0.000$ & $-0.003$ & $\mathbf{-0.299}$ & $+0.000$ & $-0.001$ & $-0.015$ \\
\texttt{thgl-forum}      & $-0.001$ & $\mathbf{-0.004}$ & $+0.000$ & $+0.001$ & $+0.000$ & $+0.000$ \\
\texttt{thgl-github}     & $+0.000$ & $-0.002$ & $-0.041$ & $+0.000$ & $+0.002$ & $\mathbf{-0.055}$ \\
\texttt{thgl-myket}      & $-0.001$ & $\mathbf{-0.020}$ & $-0.000$ & $+0.000$ & $-0.001$ & $+0.001$ \\
\midrule
\multicolumn{7}{@{}l}{\textit{Temporal Knowledge Graphs}} \\
\texttt{tkgl-smallpedia} & $-0.009$ & $+0.019$ & $-0.002$ & $\mathbf{-0.174}$ & $-0.003$ & $-0.003$ \\
\texttt{tkgl-polecat}    & $-0.004$ & $+0.044$ & $-0.008$ & $-0.009$ & $-0.008$ & $\mathbf{-0.014}$ \\
\texttt{tkgl-icews}      & $\mathbf{-0.047}$ & $+0.026$ & $+0.009$ & $-0.008$ & $-0.004$ & $-0.006$ \\
\texttt{tkgl-wikidata}   & $+0.005$ & $-0.024$ & $-0.007$ & $\mathbf{-0.029}$ & $-0.002$ & $+0.000$ \\
\bottomrule
\end{tabular}
\end{table}

\paragraph{Bank-scope drop: the bank does multi-scale work where it matters.}

Table~\ref{tab:abl-bank} zeros each bank scope's three columns individually, leaving the six base features and the other two bank scopes intact. On the THGs dropping any single scope is near-neutral (all $|\Delta| \le 0.006$): the three timescale families carry redundant signal, and the bank degrades gracefully. On the collapsed year-scale TKGs the exact-triple scope is load-bearing: removing it costs \texttt{tkgl-smallpedia} $-0.082$ and \texttt{tkgl-wikidata} $-0.010$, the largest single-scope decrements on those datasets.

\begin{table}[!ht]
\centering
\caption{Bank-scope drop (test MRR $\Delta$ vs.\ full model, seed 1337). Each entry zeros that scope's three bank columns, keeping base features and the other scopes. Largest drop per row in \textbf{bold}.}
\label{tab:abl-bank}
\footnotesize
\setlength{\tabcolsep}{6pt}
\begin{tabular}{lccc}
\toprule
Dataset & drop srd & drop rd & drop d \\
\midrule
\multicolumn{4}{@{}l}{\textit{Temporal Heterogeneous Graphs}} \\
\texttt{thgl-software}   & $-0.0001$ & $\mathbf{-0.0009}$ & $-0.0003$ \\
\texttt{thgl-forum}      & $-0.0009$ & $\mathbf{-0.0057}$ & $-0.0042$ \\
\texttt{thgl-github}     & $+0.0000$ & $+0.0048$ & $\mathbf{-0.0029}$ \\
\texttt{thgl-myket}      & $+0.0008$ & $\mathbf{-0.0026}$ & $+0.0006$ \\
\midrule
\multicolumn{4}{@{}l}{\textit{Temporal Knowledge Graphs}} \\
\texttt{tkgl-smallpedia} & $\mathbf{-0.0823}$ & $-0.0030$ & $+0.0022$ \\
\texttt{tkgl-polecat}    & $-0.0001$ & $+0.0001$ & $\mathbf{-0.0028}$ \\
\texttt{tkgl-icews}      & $\mathbf{-0.0049}$ & $-0.0014$ & $-0.0004$ \\
\texttt{tkgl-wikidata}   & $\mathbf{-0.0102}$ & $-0.0044$ & $+0.0003$ \\
\bottomrule
\end{tabular}
\end{table}

\paragraph{Calibration scheme: geometric vs.\ median and quantile.} The preceding ablations isolate \emph{which} features and scopes matter; this one isolates \emph{how the bank's decay timescales are chosen}. The deployed median-anchored geometric spacing at $\gamma=2$ is compared against two alternatives. Both reuse the identical pipeline and differ only in how each scope's decay timescales are set. The \emph{median} variant uses $n=1$: a single median-anchored decay column per scope. The \emph{quantile} variant also uses $n=3$, but reads the three half-lives at the $0.25/0.50/0.90$ quantiles of each scope's gap distribution rather than placing them geometrically about the median. All three schemes are run on all eight datasets at the full five seeds; Table~\ref{tab:abl-calib} reports the result, grouped by gap regime.

Geometric calibration is best on both concentrated year-scale TKGs. It wins by $+0.0152$ over the better alternative on \texttt{tkgl-smallpedia} and $+0.0098$ on \texttt{tkgl-wikidata}, margins several times the seed standard deviation. The mechanism is the one analyzed in \S\ref{sec:smallpedia}. On the year-scale graphs the inter-recurrence gaps concentrate near one year. The quantile scheme then reads three nearly-identical half-lives, and the median scheme has only one by construction. Both therefore collapse toward a single effective decay. Median-anchored geometric spacing avoids this degeneracy by forcing $\{0.5,1,2\}$-year separation. On the remaining six datasets the gaps are spread across their natural range, and the three schemes agree to within seed noise ($|\Delta| \le 0.0015$). The geometric scheme's advantage is thus specific to the collapsed year-scale regime. It costs nothing where multi-timescale separation is unnecessary.

\begin{table}[t]
\centering
\caption{Calibration-scheme ablation (test MRR, mean $\pm$ unbiased standard deviation over seeds $\{1337,\dots,1341\}$, $n=5$). The \emph{gap regime} column is the dataset's inter-recurrence gap scale and whether gaps are concentrated or spread. \emph{Median}: one median-anchored decay column per scope ($n=1$). \emph{Quantile}: half-lives at the $0.25/0.50/0.90$ gap quantiles. \emph{Geometric}: the deployed scheme, three half-lives placed geometrically at $\{m/2, m, 2m\}$ around the median ($\gamma=2$). The last column is geometric minus the stronger alternative. Rows group the two concentrated year-scale TKGs (top) and the six spread-gap datasets (bottom).}
\label{tab:abl-calib}
\footnotesize
\setlength{\tabcolsep}{5pt}
\resizebox{\textwidth}{!}{%
\begin{tabular}{llrrrr}
\toprule
Dataset & Gap regime & \multicolumn{1}{l}{Median} & \multicolumn{1}{l}{Quantile} & \multicolumn{1}{l}{Geometric (deployed)} & \multicolumn{1}{l}{$\Delta$ vs.\ best alt.} \\
\midrule
\multicolumn{6}{@{}l}{\textit{Year-scale (concentrated gaps)}} \\
\texttt{tkgl-smallpedia} & year (concentrated)    & $0.5854 \pm 0.0066$ & $0.5984 \pm 0.0025$ & $\mathbf{0.6136 \pm 0.0017}$ & $+0.0152$ \\
\texttt{tkgl-wikidata}   & year (concentrated)    & $0.6299 \pm 0.0142$ & $0.6203 \pm 0.0201$ & $\mathbf{0.6397 \pm 0.0082}$ & $+0.0098$ \\
\midrule
\multicolumn{6}{@{}l}{\textit{Spread-gap}} \\
\texttt{thgl-forum}      & seconds (spread)       & $0.7285 \pm 0.0027$ & $\mathbf{0.7321 \pm 0.0023}$ & $0.7306 \pm 0.0021$ & $-0.0015$ \\
\texttt{thgl-github}     & seconds (spread)       & $\mathbf{0.7855 \pm 0.0004}$ & $\mathbf{0.7855 \pm 0.0004}$ & $0.7842 \pm 0.0008$ & $-0.0013$ \\
\texttt{thgl-software}   & minutes (spread)       & $0.5011 \pm 0.0009$ & $\mathbf{0.5014 \pm 0.0007}$ & $0.5013 \pm 0.0007$ & $-0.0001$ \\
\texttt{thgl-myket}      & minutes (spread)       & $0.9050 \pm 0.0011$ & $0.9049 \pm 0.0009$ & $\mathbf{0.9056 \pm 0.0004}$ & $+0.0006$ \\
\texttt{tkgl-polecat}    & day (spread)           & $\mathbf{0.1715 \pm 0.0045}$ & $0.1695 \pm 0.0022$ & $0.1707 \pm 0.0046$ & $-0.0008$ \\
\texttt{tkgl-icews}      & day (spread)           & $0.2667 \pm 0.0047$ & $0.2692 \pm 0.0053$ & $\mathbf{0.2694 \pm 0.0034}$ & $+0.0002$ \\
\bottomrule
\end{tabular}%
}
\end{table}

\paragraph{Spacing-factor and timescale-count sensitivity.} The spacing factor $\gamma$ and the timescale count $n$ are the only hand-set calibration hyperparameters (\S\ref{sec:bank}). The scheme ablation above varies how timescales are chosen. Here each is varied directly, one axis at a time through the deployed $(\gamma=2, n=3)$ point, to test whether either is finely tuned. Both sweeps use five seeds on the two gap-concentrated datasets where the calibration hyperparameters are load-bearing. On the six spread-gap datasets both are inconsequential.

Table~\ref{tab:abl-phi} reports how test MRR varies with $\gamma$ and $n$ around the deployed $(\gamma=2, n=3)$ point. Both are stable. The deployed $\gamma=2$ is best within seed noise, and the benefit saturates at $n=3$, with larger counts adding nothing. The deployed $(\gamma=2, n=3)$ is therefore not finely tuned, and $n=3$ is the smallest count capturing the full benefit.

\begin{table}[htbp]
\centering
\caption{Calibration hyperparameter sensitivity on the two gap-concentrated year-scale TKGs (test MRR, five seeds), as two orthogonal slices through the deployed $(\gamma=2, n=3)$ point. \emph{Top:} spacing factor $\gamma$ at fixed $n=3$. \emph{Bottom:} timescale count $n$ at fixed $\gamma=2$. Best per row in bold; the $(\gamma=2, n=3)$ column is the main-results run (Table~\ref{tab:main}).}
\label{tab:abl-phi}
\footnotesize
\setlength{\tabcolsep}{8pt}
\renewcommand{\arraystretch}{0.9}
\resizebox{\textwidth}{!}{%
\begin{tabular}{lcccc}
\toprule
\multicolumn{5}{l}{\emph{Spacing factor $\gamma$ ($n=3$ fixed)}} \\
Dataset & $\gamma=1.5$ & $\gamma=2$ (deployed) & $\gamma=3$ & $\gamma=4$ \\
\midrule
\texttt{tkgl-smallpedia} & $0.6045 \pm 0.0058$ & $\mathbf{0.6136 \pm 0.0017}$ & $0.6119 \pm 0.0028$ & $0.6097 \pm 0.0023$ \\
\texttt{tkgl-wikidata}   & $0.6368 \pm 0.0059$ & $\mathbf{0.6397 \pm 0.0082}$ & $0.6348 \pm 0.0162$ & $0.6331 \pm 0.0156$ \\
\midrule
\multicolumn{5}{l}{\emph{Timescale count $n$ ($\gamma=2$ fixed)}} \\
Dataset & $n=2$ & $n=3$ (deployed) & $n=4$ & $n=5$ \\
\midrule
\texttt{tkgl-smallpedia} & $0.5997 \pm 0.0056$ & $0.6136 \pm 0.0017$ & $\mathbf{0.6141 \pm 0.0029}$ & $0.6133 \pm 0.0027$ \\
\texttt{tkgl-wikidata}   & $0.6385 \pm 0.0079$ & $\mathbf{0.6397 \pm 0.0082}$ & $0.6391 \pm 0.0077$ & $0.6328 \pm 0.0137$ \\
\bottomrule
\end{tabular}%
}
\end{table}

\paragraph{Relation-awareness factorial: decomposing the value of the relation.}
\label{app:relation-ablation}
\rev{This section varies relation-awareness itself, decomposing it into the two components that can be manipulated independently. Where \S\ref{sec:relation-value} asks how much recurrence structure the relation exposes in the data, this asks how much test MRR the model gains from using it.

Relation-awareness enters the model at two points. The \emph{scope} factor conditions the memorization features, collapsing the relation-conditioned scopes $(s,r,d)$ and $(r,d)$ to their relation-agnostic counterparts $(s,d)$ and $(d)$. The \emph{weight} factor indexes the parameters, replacing the per-relation matrix $\theta \in \mathbb{R}^{|\mathcal{R}| \times 15}$ with a single shared vector $\theta \in \mathbb{R}^{15}$. Bank recalibration is nested within the scope factor rather than being a third free factor. Half-lives are calibrated from each scope's own gaps (Eq.~\ref{eq:geometric-cal}), so collapsing the scopes forces recalibration. Otherwise the collapsed cells would carry half-lives fitted to scopes they can no longer read. Crossing the two factors gives four cells: (A) the full model, (B) relation-conditioned scopes with shared weights, (C) collapsed scopes with per-relation weights, and (D) the fully relation-blind model. Cell (B) is exactly a conditional-logit fit of the $15$ features, that is, a logistic regression over the same tabular representation. All cells share the pipeline, negative sampler, and selector of \S\ref{sec:setup}, and differ only in the two switches.}

\begin{table}[t]
\centering
\begingroup\revon
\caption{\rev{Relation-awareness factorial (test MRR, five-seed means, seeds $1337$--$1341$). (A) full model, from Table~\ref{tab:main}. (B) relation-conditioned scopes, shared weights, equivalently logistic regression on the $15$ features. (C) collapsed scopes, per-relation weights. (D) relation-blind. $\Delta_{A-B}$ isolates per-relation weighting, $\Delta_{A-C}$ relation-conditioned features, and $\Delta_{A-D}$ the total. Retention $\mathrm{ret}$ is from Table~\ref{tab:recurrence}. Entries marked $\ast$ are unstable under the val-best-epoch selector and are not point estimates, as are the $\Delta$ values computed from them.}}
\label{tab:relation-ablation}
\footnotesize
\setlength{\tabcolsep}{2.5pt}
\begin{tabular}{l|c|cccc|ccc}
\toprule
Dataset & $\mathrm{ret}$ & (A) Full & (B) Shared & (C) Collapsed & (D) Blind & $\Delta_{A-B}$ & $\Delta_{A-C}$ & $\Delta_{A-D}$ \\
\midrule
\multicolumn{9}{@{}l}{\textit{Temporal Heterogeneous Graphs}} \\
\texttt{thgl-software}   & 0.68 & 0.501 & 0.468 & 0.492 & 0.466 & $+0.033$ & $+0.009$ & $+0.035$ \\
\texttt{thgl-forum}      & 1.00 & 0.731 & 0.736 & 0.730 & 0.736 & $-0.005$ & $+0.001$ & $-0.005$ \\
\texttt{thgl-github}     & 0.84 & 0.784 & 0.679 & 0.783 & 0.701 & $+0.105$ & $+0.002$ & $+0.083$ \\
\texttt{thgl-myket}      & 1.00 & 0.906 & 0.907 & 0.903 & 0.903 & $-0.001$ & $+0.003$ & $+0.003$ \\
\midrule
\multicolumn{9}{@{}l}{\textit{Temporal Knowledge Graphs}} \\
\texttt{tkgl-smallpedia} & 0.95 & 0.614 & 0.574 & 0.389 & 0.581 & $+0.040$ & $+0.225$ & $+0.033$ \\
\texttt{tkgl-polecat}    & 0.89 & 0.171 & 0.180$^\ast$ & 0.177$^\ast$ & 0.267$^\ast$ & $-0.009^\ast$ & $-0.006^\ast$ & $-0.096^\ast$ \\
\texttt{tkgl-icews}      & 0.54 & 0.269 & 0.277$^\ast$ & 0.168 & 0.212$^\ast$ & $-0.008^\ast$ & $+0.101$ & $+0.057^\ast$ \\
\texttt{tkgl-wikidata}   & 0.94 & 0.640 & 0.633 & 0.636 & 0.636 & $+0.007$ & $+0.004$ & $+0.004$ \\
\bottomrule
\end{tabular}
\endgroup
\end{table}

\rev{Two readings follow. First, removing relation-awareness entirely costs little, with $\Delta_{A-D}$ ranging from $-0.005$ to $+0.083$ across the six datasets with stable estimates. That total understates the individual factors, which interact strongly on \texttt{tkgl-smallpedia}. On \texttt{thgl-forum} the blind model is marginally ahead, as the retention analysis of \S\ref{sec:relation-value} predicts for a dataset whose destinations recur under a single relation. The \texttt{tkgl-polecat} lead of $0.096$ comes from a selector-unstable cell. Second, the contribution is dataset-specific, tracking neither relation count nor retention.}

\rev{The two factors do not decompose additively. On \texttt{thgl-github}, $\Delta_{A-B}$ and $\Delta_{A-C}$ sum to $0.023$ more than $\Delta_{A-D}$, and on \texttt{tkgl-smallpedia} to $0.232$ more. The cells are therefore reported individually rather than as shares of a total. The \texttt{tkgl-smallpedia} interaction has a specific cause. Cell (C) supplies $566$ weight vectors over features that carry no relation signal, fitting the $K=20$ in-pool training distribution at the expense of the $47{,}433$-candidate evaluation pool (\S\ref{sec:polecat-icews}). It accordingly falls below cell (D) there, and carries the largest seed spread in the table.}

\rev{The scope factor is not confounded with feature starvation. Collapsing makes the $srd$ slot equal the $sd$ slot and the $rd$ bank equal the $d$ bank, leaving $10$ distinct features of the $15$, and because the scorer is linear, collinear columns do not change achievable performance. The weight factor does reduce capacity, from $15|\mathcal{R}|$ parameters to $15$, so it is varied on its own axis. Entries marked $\ast$ are the exception. On \texttt{tkgl-polecat}, and in the shared-weight and blind cells on \texttt{tkgl-icews}, the selected epoch varies across seeds from the first to the last, so the smoothing window of \S\ref{sec:selector} does not stabilize the choice. The collapsed-scope cell on \texttt{tkgl-icews} does not, with a seed spread of $0.0020$, nor does the deployed configuration (Table~\ref{tab:main}).

None of the structural properties available before running the model predicts whether relation-awareness will help. Relation count, retention, and negative-pool size all fail across these eight datasets, so the contribution has to be measured rather than anticipated.}

\end{document}